\documentclass[12pt, a4paper]{article}

\usepackage[margin=1in]{geometry}
\usepackage{times}
\usepackage{booktabs}

\usepackage{graphicx}
\usepackage{float}
\usepackage{amssymb}
\usepackage{amsmath}
\usepackage{longtable}
\usepackage[table]{xcolor} 
\usepackage[round]{natbib}
\usepackage{lineno}

\usepackage{hyperref} 
\hypersetup{
    colorlinks=true,   
    linkcolor=blue,    
    citecolor=blue,    
    urlcolor=blue      
}

\let\cite\citep

\usepackage{authblk}

\title{Generative AI for Visualizing Highway Construction Hazards Through Synthetic Images and Temporal Sequences}
\author[1]{Trevor Neece}
\author[1]{Mason Smetana}
\author[1]{Lev Khazanovich}

\affil[1]{University of Pittsburgh, Pittsburgh, PA, USA}

\date{}

\makeatletter
\AtBeginDocument{%
  \@ifundefined{DIFaddbeginFL}{}{}%
  \@ifundefined{DIFaddendFL}{}{}%
  \@ifundefined{DIFdelbeginFL}{}{}%
  \@ifundefined{DIFdelendFL}{}{}%
}
\makeatother

\begin{document}

\maketitle
\begin{center}
\colorbox{black!5}{
    \parbox{0.9\textwidth}{
        \centering
        \small
        \textit{This is the final accepted version of a manuscript published in Construction Innovation.}\\
        \vspace{0.5em}
        The final published Version of Record can be found at:\\
        \url{https://doi.org/10.1108/CI-04-2026-0175}
    }
}
\end{center}
\begin{abstract} 
\noindent \textbf{Purpose:} Highway construction workers face a high risk of serious injury or death. Image-based training materials depicting hazardous scenarios are essential for engaging safety instruction but remain scarce due to ethical and logistical barriers. This study develops and evaluates a generative AI methodology for producing synthetic visualizations of highway construction hazards from OSHA Severe Injury Report narratives.

\noindent \textbf{Design/methodology/approach:} Two modes were developed: a single-pass approach yielding one image per incident, and a temporal approach producing a four-stage sequence. Both approaches were evaluated using CLIP-based semantic retrieval and assessed by domain experts across dimensions such as educational utility, fidelity, and alignment.

\noindent \textbf{Findings:} Single-pass images achieved 81.1\% educational acceptability with fidelity and alignment scores of 4.14/5 and 4.07/5. Temporal sequences achieved 61.0\% acceptability with comparable alignment (3.94/5) but lower fidelity (3.51/5). CLIP-based retrieval revealed that both modes produce images with statistically significant retrieval capabilities.

\noindent \textbf{Originality:} This is among the first studies to leverage modern autoregressive image generation models for visualizing construction hazards from reported severe injuries and to generate temporally sequenced hazard imagery. A new multi-dimensional evaluation framework was also developed to support future research in this domain.

\noindent \textbf{Practical implications:} This work enables safety trainers to pair narrative storytelling with visual learning material without photographing real-world hazards. The new data can be integrated into existing construction safety training systems such as toolbox talks, VR/AR platforms, or OSHA educational programs.
\end{abstract}

\vspace{1em}

\section{Introduction}
\label{sec:introduction}
Workers in highway construction work zones are at very high risk of serious injury or death, with recent statistics showing that work zone fatalities increased by approximately 21.5\% between 2015 and 2023~\cite{wzsic_data}. Within a work zone, construction labor occurs alongside live traffic, leading to interactions between workers and vehicles and thereby increasing the risk of injury for both drivers~\cite{weng2011analysis} and workers~\cite{nnaji2020improving}. This labor is vital to the development of sustainable built environments and to overall societal prosperity, as road and highway construction significantly impacts economic development~\cite{gkritza2008influence}. However, persistent safety risks indicate that the industry needs to place greater emphasis on safety-related research~\cite{eseonu2018reducing}.

Workers tend to act based on their perception of risk, which is often misaligned with the true risk~\cite{asnar2008perceived}. As such, calibrating a worker's perceived risk with the objective risk of a highway construction task is a key driver of safety~\cite{Wilkins2011Construction}. Safety knowledge, therefore, is highly correlated with safety performance~\cite{Sawacha1999Factors}. Although health and safety training is common in construction, a major limitation is the overuse of ineffective presentation techniques such as text-based lectures. When high-engagement training, characterized by interactive and visually centered learning activities, is employed, skill training and safety knowledge acquisition are up to three times as effective as low-engagement methods~\cite{burke2006relative}. Given the elevated risks in work zones, safety instructors must be keenly aware of which methods keep learners engaged to ensure that vital safety knowledge is effectively communicated.

The effectiveness of visually centered instruction is based on established theories of multimedia learning. Paivio's dual coding theory posits that information learned through both verbal and visual channels is retained more readily than information presented through a single channel~\cite{paivio1986mental}. Mayer's cognitive theory of multimedia learning extended this, showing that words paired with relevant images promote deeper learning than words alone~\cite{mayer2002multimedia}. Relatedly, in a lecture or presentation setting, audience attention typically alternates between engaged and unengaged in cycles that often grow shorter as the session progresses~\cite{bunce2010long}, and visually centered, high-engagement training has been shown to counteract these diminishing cycles~\cite{albert2020developing}. 

With regard to safety training, techniques such as narrative-based training, in which learners observe a construction hazard evolving over time and engage in discussion, can further enhance engagement and improve retention of safety knowledge~\cite{eggerth2018evaluation}. Safety-based storytelling, in which a safe construction scene transitions to a critical hazard, provides this narrative structure, giving learners the contextual buildup needed to recognize warning signs before a hazard peaks~\cite{cullen2008tell}. Framing a hazard as an unfolding sequence also aligns with hazard communication principles, which hold that effective warnings convey not only the hazard itself but the precursor conditions and behaviors that precede it~\cite{wogalter2006handbook}. However, these techniques are often hindered by a fundamental gap in data availability: there are very few high-quality image datasets of construction hazards~\cite{guo2021computer}. Capturing such imagery directly is constrained by site access limitations, the ethical problem that authentic photographs would document workers engaged in unsafe practices that should be interrupted rather than recorded, and companies' reluctance to release any such imagery publicly. 

Although image data are scarce, large databases of text-based descriptions of construction hazards exist. For example, since 2015, OSHA has required U.S. employers to report all severe occupational injuries, including location, incident description, and injury details~\cite{gomes2023time}. With advanced image-generation capabilities enabled by methods such as Stable Diffusion and generative artificial intelligence (AI), such text-based descriptions can be transformed into images~\cite{acharjee2025data}. By leveraging generative models, the industry can bypass the ethical and logistical hurdles of capturing real-world images of hazards. Consequently, these synthetic visuals can be integrated into the high-engagement, narrative-based training modules previously discussed, providing workers with vivid, contextually relevant training material. Synthetic images would address many data-availability gaps and provide a scalable framework for calibrating workers' risk perception, potentially reducing fatalities.

Despite the promise of generative AI, stochastic models exhibit substantial output variance. In some applications, this variability can be modulated via a "temperature" parameter; however, for tasks that require creativity, such as image generation, such control is not always feasible or desirable~\cite{li2025exploring}. A single-attempt image generation will often produce a pedagogically incorrect image on the first iteration, as text-to-image models frequently struggle with fine-grained semantic alignment and complex spatial reasoning in a single pass~\cite{khan2025test}. Instead, achieving reliable output typically requires an iterative process in which initial generations are evaluated and refined~\cite{jeon2025iterative}. This study asks how generative AI can transform OSHA incident narratives into realistic, semantically valid imagery of construction hazards that evolves over time, and how such imagery should be evaluated for educational use. The following section situates this work within the broader literature on safety training, synthetic image generation in construction, and the evaluation of AI-generated imagery, and identifies the gaps this study aims to address.

\subsection{Literature Review}
\label{sec:literaturereview}
Building on the engagement and data-scarcity issues outlined above, prior work has further established that high-engagement methods are especially effective for high-exposure occupations such as highway construction~\cite{burke2011dread}, and that interactive multimedia outperforms lecture-based training while being cost-effective to redeploy across sessions~\cite{cherrett2009making, albert2021designing}. The corresponding shortage of hazard imagery has multiple drivers: expert annotation is expensive~\cite{ricketts2023scoping}, site cameras are positioned for coverage rather than detail~\cite{tian2024review}, and companies withhold imagery for liability reasons~\cite{tanga2022data}. Methods for generating project-specific hazard visualizations are therefore needed to fill this gap.

Accordingly, the generation of synthetic images for diverse datasets containing hazardous scenarios is a recently emerging research topic. \citet{lee2022synthetic} leveraged game-engine renderings of construction machines, imposed on synthetic backgrounds, to create a new dataset of heavy-machine operations. \citet{xiong2021machine} used game-engine-generated dust effects, applied only to construction images, for dataset augmentation, achieving an F1 score of 0.93 when a YOLO-v3 model trained exclusively on 3,860 synthetic images was evaluated on 1,015 real-world construction photographs. \citet{wei2021synthetic} developed a BIM-to-game-engine framework for generating synthetic images of construction projects using digital ground truth. \citet{kim2024image} used text-to-image generative models to create a dataset of 3,585 hazardous images spanning 27 hazard categories, and then trained a deep neural network to perform hazard recognition on images, achieving a mean average precision of approximately 64\% for object detection and 60\% for segmentation.  \citet{neece2026comparative} used game-engine-based images of construction hazards from construction health and safety training modules as benchmarking data for AI-based image hazard identification, reporting that 95\% confidence intervals for model recognition scores substantially overlapped between real and synthetic image classes across all three multimodal AI models evaluated, indicating statistically similar performance. Collectively, these studies demonstrate that synthetically generated construction imagery, produced via game engines, BIM pipelines, or generative AI, yields performance competitive with real-world data. Notably, the studies above employ synthetic imagery primarily as training or evaluation data for machine perception models rather than as instructional content for human learners. The use of generative AI to produce educational visualizations for safety training remains comparatively underexplored.

Image synthesis using generative AI is dominated by two model architectures. The first is diffusion models, which generate an image by iteratively denoising a random field until a coherent output emerges~\cite{blattmann2023align}. The second is autoregressive models, which generate an image as a sequence of tokens, analogous to how large language models produce text~\cite{zuo2025nano}. A key advantage of autoregressive models is their image-to-image capabilities without degrading the source. If a diffusion-based model is used to edit an image, the entire image will be resynthesized, leaving all regions subject to stochastic variation. Open-source image generation remains predominantly diffusion-based and has not yet matched the region-specific editing fidelity of closed frontier systems such as Google's Gemini 3 Pro~\cite{googledeepMind2025nanobanana}, which pairs autoregressive generation with multimodal reasoning. 

Generative models have demonstrated capabilities in text-to-image synthesis that bypass traditional data collection barriers. \citet{Hong_Choi_Ham_Jeon_Kim_2024} used a stable diffusion model (Diffusion SDXL 1.0 with the HuggingFace diffusers library) to generate images representing construction tasks and found that 82.01\% of the synthetic images were suitable representations of the desired task. Additionally, a CNN model trained on synthetic images with and without context-based labels saw classification accuracy of 89.09\% and 86.51\%, respectively. \citet{acharjee2025data} generated images of general workplace-specific accident scenarios using accident narratives from OSHA's Severe Injury Reports (SIRs) with Stable Diffusion 3.5 Large. The workflow involved large language models (LLMs) to generate descriptions of accident scenarios and a pretrained diffusion model to create new image-label pairs. The methodology was evaluated through manual inspection, which found that 86\% of images accurately represented the scenario described in the corresponding prompts. This work acknowledges a limitation of diffusion: although the generated images capture the intended hazards, they often appear overly artistic rather than photorealistic. Along with improved fidelity, new autoregressive models can perform image-to-image translations without degrading the original input~\cite{xiao2025deterministic}.

Recent frontier multimodal LLMs, such as Google's Gemini 3 Pro~\cite{googledeepMind2025nanobanana}, have demonstrated region-specific editing capabilities, enabling localized semantic modifications to target image regions while preserving the structural and appearance fidelity of non-target content. As such, consistent image-to-image generation enables the creation of sequential images that depict how hazards evolve, without degrading vital image components, such as construction tasks and highway infrastructure. However, these models have not yet been evaluated for Construction 4.0-related applications, nor has the image-to-image generation pipeline been developed. 

While generative AI shows significant promise in creating image-based visualizations that do not yet exist, vision-language models (VLMs) exhibit known limitations in spatial reasoning~\cite{kamath2023s} and domain-specific accuracy in construction~\cite{adil2025using}, which must be addressed for reliable application in safety-critical training contexts. \citet{kamath2023s} evaluated 18 vision-language models on spatial reasoning tasks (i.e., reliably distinguish between "right" vs "left") and found that all performed poorly. \citet{neece2026comparative} demonstrated that VLMs have a baseline capability for hazard recognition, but perform worse in identifying electrocution and struck-by hazards.

To evaluate synthetic imagery, \citet{otani2023toward} found that most studies use automated evaluation metrics to validate generated images or rely on human evaluation, which lacks repeatability. Additionally, their work defined best practices for evaluating AI-generated images, specifically a 1-5 Likert-scale human grade for fidelity and alignment. \citet{cook2024ticking} proposed a checklist-based evaluation methodology that increased agreement between LLM-based and human-based evaluations of generated content. \citet{lu2025generating} generated synthetic construction images of construction machinery and validated the new dataset using a quantitative, computer-vision-based recall assessment and a qualitative, expert-based Turing Test. Automated evaluation of AI-generated imagery generally relies on an embedding model that converts images and text descriptions into vectors in a shared numerical space, allowing their similarity to be measured by distance. The most widely used procedure is Contrastive Language-Image Pre-training (CLIP)~\cite{radford2021learning}, though other methods exist for specialized applications, such as \citet{zhang2024long}'s Long-CLIP, which is specialized for longer text inputs, and \citet{lin2024evaluating}'s model for visual question answering. In the highway construction domain, non-automated expert-based evaluation criteria for safety-related applications remain underdeveloped~\cite{neece2026immersive}.

The studies reviewed above establish that synthetic imagery can serve as a substitute for photographs for construction safety. However, training applications require both technical validity and educational utility. High-engagement visual training significantly outperforms lecture-based instruction in both knowledge acquisition and retention~\cite{burke2006relative}, and narrative-based techniques, in which learners observe a hazard visually unfolding over time, further enhance engagement and the practical application of safety knowledge~\cite{cullen2008tell, eggerth2018evaluation}. A static image of a hazard does not maximally fulfill this narrative function. Ideally, multiple images would depict how hazards evolve over time, providing narrative cues for workers to learn from. Despite growing interest in AI-generated imagery for construction, no prior work has examined the generation of temporally sequenced hazard imagery that depicts the progressive escalation from safe conditions to critical safety violations. Additionally, highly realistic safety-based images have not yet been evaluated using an expert-based framework in the highway construction domain.

\subsection{Research Aims and Objectives}
There is a clear gap in the availability of image-based safety training materials depicting hazardous highway construction scenarios. While OSHA's databases provide a rich corpus of historical incident narratives, no established methodology systematically transforms these records into visual training assets. Concurrently, recent advances in frontier image generation models have demonstrated region-specific editing capabilities and high-fidelity image-to-image translation that were not previously achievable, creating a new technical opportunity for synthetic safety visualization. If these hazard visualization images can be created, the new data can be integrated into existing construction safety training systems such as toolbox talks, VR/AR platforms, or OSHA educational programs. This study, therefore, addresses the following research question: How can generative AI be leveraged to produce realistic, structurally coherent, and semantically valid synthetic imagery of construction safety hazards that evolve over time, and what methodologies should guide the evaluation of such imagery? 

\begin{figure}[H]
    \centering
    \includegraphics[width=0.9\textwidth]{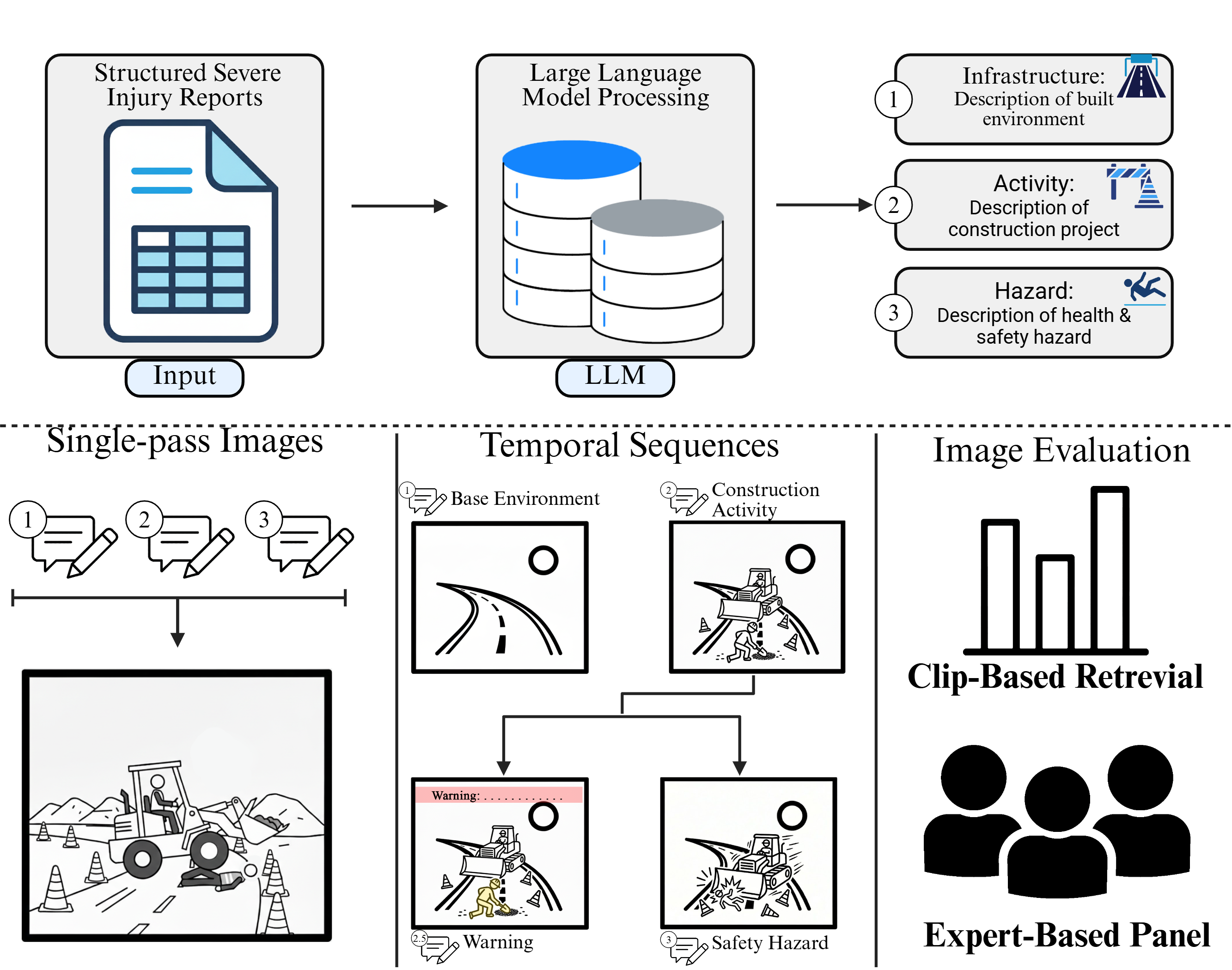}
    \caption{Graphical overview of the proposed contributions of this study.}
    \label{fig:fig1}
\end{figure}

To address this research question, the study pursues three objectives, illustrated in Figure~\ref{fig:fig1}:
\begin{enumerate}
    \item A framework for single-pass generation of highway construction hazard imagery using structured prompt engineering, in which a layered prompting strategy decomposes complex hazard scenes into infrastructure, activity, and hazard elements to produce photorealistic imagery from a single text-to-image generation pass.
    \item A framework for generating temporal image sequences that depict hazard progression, from safe configurations through early-warning states to critical violations, using iterative image-to-image conditioning to maintain visual continuity across sequential frames. 
    \item A framework for both automated and manual evaluation of synthetic construction safety imagery, addressing dimensions such as fidelity, alignment, educational utility, and retrieval. 
\end{enumerate}

\section{Methodology}
\begin{figure}[H]
    \centering
    \includegraphics[width=0.9\textwidth]{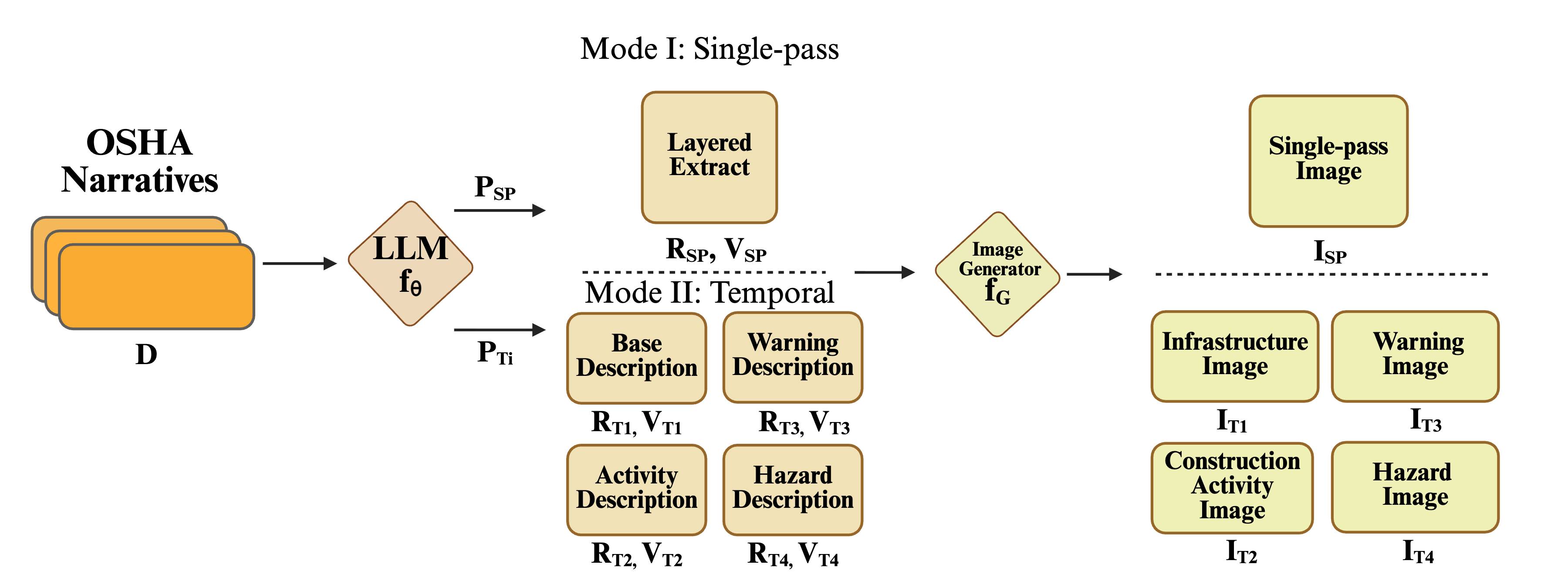}
    \caption{Overview of the proposed methodology, illustrating the flow from OSHA SIR narratives through LLM-based scene description generation to two image generation pipelines (Mode I: Single-pass and Mode II: Temporal).}
    \label{fig:math}
\end{figure}

The proposed methodology is illustrated in Figure~\ref{fig:math}. OSHA SIRs serve as the primary input, processed by an LLM $f_\theta$ to produce structured scene descriptions that are then fed into two distinct image-generation pipelines. In Mode I, a single-pass layered extract drives a one-pass generation through the image generator $f_G$, producing $I_{SP}$. In Mode II, four sequential descriptions ($R_{T1}$, $R_{T2}$, $R_{T3}$, and $R_{T4}$) drive an iterative image-to-image conditioning pipeline, yielding the temporal sequence $\{I_{T1}, I_{T2}, I_{T3}, I_{T4}\}$. The resulting images are then evaluated using two complementary strategies: CLIP-based retrieval and an expert-based panel, as described in Section~\ref{sec:eval}.

\subsection{Severe Injury Reports for Highway Construction}
\label{sec:sir}
The OSHA SIRs were used as the ground-truth data source for generating images. This database was selected for its completeness and text-rich abstracts that describe the reported severe injury. The OSHA SIR database, spanning 2015–2021, contains more than 70,000 records across all North American Industry Classification System (NAICS) industry codes. For this study, only NAICS Code 237310 was used, which corresponds to Highway, Street, and Bridge Construction. After filtering, 1198 structured incidents were collected for potential inclusion.

Each incident record $D$ drawn from the narratives dataset is treated as a structured document containing 26 fields:
$$
D = \{\text{site\_address}, \text{abstract\_text}, \text{event\_keyword}, \text{src\_of\_injury}, \text{event\_type}, \ldots\}
$$
While fields such as "\text{abstract\_text}", "\text{event\_keyword}", and "\text{src\_of\_injury}" carry the majority of the scene-relevant information, administrative fields such as "\text{activity\_nr}", "\text{reporting\_id}", and "\text{summary\_nr}" are included on the basis that their presence does not degrade generation quality, and that manual selection of fields would introduce unnecessary subjective choice. \citet{smetana2024highway, smetana2026improving} previously analyzed this dataset using LLMs across the same NAICS 237310 subset, identifying the free-text investigator narrative (field "abstract\_text") as the primary source of analytical value and validating its semantic richness for downstream LLM tasks.

To evaluate semantic alignment between generated images and their source descriptions, a random sample of 75 SIR records was drawn from this dataset. Based on the "event\_keyword" field, the sample was dominated by Struck-By incidents ($n=37$, 49.3\%), followed by Fall from Elevation ($n=10$, 13.3\%), Caught in/Between ($n=9$, 12.0\%), Other ($n=7$, 9.3\%), and Struck Against ($n=6$, 8.0\%), with five smaller categories accounting for the remaining six records (8.0\%). To confirm that this is a representative sample of records, it was compared to the true population distribution (Struck-By 51.8\%, Caught in/Between 17.9\%, Fall from Elevation 12.9\%). The sampled population closely mirrors the full population, and a chi-square goodness-of-fit test found no significant difference between the two distributions ($\chi^2 = 6.32$, $df = 5$, $p = 0.276$).

For each of the 75 selected records, two independent image-generation iterations were performed under both modes described in Sections~\ref{sec:single} and~\ref{sec:temporal}. That is, for each incident record $D$, this resulted in 10 images: two single-pass hazard representations and two temporal sequences of four images each. The two iterations per methodology used identical prompts and visual constraints, differing only in the stochastic variation inherent in the image generator, allowing within-method consistency to be measured and quantifying how much semantic variability arises from generation stochasticity versus from differences in source records. The image generation was performed in February 2026.

\subsection{Generative AI Models}
This study employed two generative AI models. The text generator, denoted by $f_\theta$, was Google's Gemini 2.5 Flash~\cite{gemini25flash}, selected for its low latency and cost-efficiency in generating structured outputs. The image generator, denoted by $f_G$, was Google's Gemini 3 Pro~\cite{gemini3pro}, selected for its superior visual fidelity and region-specific editing capabilities, which are required by the temporal pipeline. The methodology can, in principle, replace either $f_\theta$ or $f_G$ with any model capable of instruction-following text generation and image-to-image generation, respectively. Models were accessed via the Google Generative AI API, with a fixed temperature of 0.0 for $f_\theta$ to minimize stochastic variation across incident records, ensuring that differences in generated output are attributable to differences in the source data $D$ rather than sampling noise. For $f_G$, the default temperature setting was used to introduce stochastic variation in image generation, as recommended in the Gemini 3 Pro documentation~\cite{googledeepMind2025nanobanana}. Both models were used without fine-tuning, with all task-specific behavior induced through prompt engineering. The complete prompts are available in the supplemental files section~A. This design choice was made due to the lack of construction-hazard imagery (see Section~\ref{sec:literaturereview}). For $f_G$, images were created at a resolution of $1408\times768$. API pricing for both models as accessed in February 2026, together with the per-image generation fee, is reported in the supplemental files section~E.

\subsection{Mode I: Single-pass Image Generation}
\label{sec:single}
This initial methodology generates an image from the raw incident record $D$ in one pass, rather than employing image-to-image scene composition. The complete record $D$ is passed to $f_\theta$ with a structured prompt $P_{SP}$ that instructs the model to define three narrative layers:
\begin{enumerate}
    \item The Infrastructure: The base roadway or work zone layout (e.g., pavement, lane markings, traffic control), with no mention of machines or workers.
    \item The Activity: The construction equipment and workers performing their routine tasks within that space.
    \item The Hazard: the peak of the event, focusing on the physical interaction between the worker and the hazard.
\end{enumerate}

The model is instructed to generate 1-2 sentences per narrative layer and to produce a cohesive response using the three descriptions, resulting in the output $R_{SP}$. Additionally, the model is instructed not to dramatize or invent information beyond what is provided in the incident report. As such, the output of the first structured prompt can be represented as:
$$
R_{SP} = f_\theta(P_{SP}(D))
$$

This description is then embedded in a visual constraints prompt $V_{SP}$. This prompt is always the same for each unique $R_{SP}$. The full text of the structured prompt and the visual constraints prompt for the single-pass mode are provided in the supplemental files sections~A.1 and~A.2. The visual constraints prompt the image generator $f_G$ to generate the single-pass image $I_{SP}$ according to a few key rules:
\begin{enumerate}
    \item The image should be a photorealistic educational visualization of the construction scenario, taken at eye level, as if the inspector were standing a few feet away.
    \item The image must be "locked" at this perspective and not panned or zoomed.
    \item The focal point (workers, vehicles, the incident, etc.) must be the sharpest and most distinct part of the image.
    \item The activity and hazard elements must be integrated seamlessly into the infrastructure, without introducing unnecessary environmental clutter.
    \item There must be no overlaid text, labels, or captions. Text must appear only on physical objects within the scene, such as on signs.
\end{enumerate}
As such, the methodology for single-pass image generation can be represented as:
$$
I_{SP} = f_G \left(  V_{SP} + R_{SP}   \right)
$$
Figure~\ref{fig:fig2} (A) presents four representative images generated via the single-pass pipeline, each produced from a distinct SIR incident record. The generated scenes depict a range of highway construction hazard types. In each case, the layered prompting strategy integrated infrastructure, activity, and hazard elements into a single coherent scene, with workers and equipment spatially positioned to reflect the described incident interaction. The first image depicts a construction vehicle crushed by a mismanaged crane load; the second and fourth images depict a worker run over by a backing vehicle; the third image depicts a roller that lost control while coasting down a hill and struck a tree. A worked single-pass example for one incident record, including the verbatim layered description $R_{SP}$ and both stochastic iterations, is provided in the supplemental files section~C (Figure~A2).

\subsection{Mode II: Temporal Sequence Generation}
\label{sec:temporal}
The temporal methodology decomposes each incident record $D$ into a set of four temporally ordered descriptions $\{R_{T1}, R_{T2}, R_{T3}, R_{T4}\}$, each generated by the language model $f_\theta$ conditioned on $D$ and prior outputs. The four stages are:
\begin{itemize}
    \item $R_{T1}$: the static base environment, including road surface, lane markings, terrain, and traffic control layout, with all workers and equipment suppressed.
    \item $R_{T2}$: the populated work zone, introducing workers and equipment into the base environment in positions consistent with the impending hazard.
    \item $R_{T3}$: an annotation that identifies the at-risk worker and includes a concise warning phrase describing the upcoming hazard, branched from $R_{T2}$.
    \item $R_{T4}$: the hazard event, depicting the worker/hazard interaction described in the source SIR record.
\end{itemize}
These descriptions then drive a sequential image-generation pipeline. The first state, $T_1$, is generated from $D$ alone:
$$
R_{T1} = f_\theta(P_{T1}(D))
$$
States 2 and 4 are generated sequentially using the previous text description and $D$:
$$
R_{T2} = f_\theta(P_{T2}(D, R_{T1}))
$$
$$
R_{T4} = f_\theta(P_{T4}(D, R_{T2}))
$$

$R_{T3}$ is also conditioned on $R_{T2}$ but instead produces an annotation specification: the identity and position of the at-risk worker to be highlighted, and a concise 5--7-word warning phrase describing the upcoming hazard. As such:
$$
R_{T3} = f_\theta(P_{T3}(D, R_{T2}))
$$
Both $R_{T3}$ and $R_{T4}$ branch from $R_{T2}$ independently:
$$
R_{T3} \leftarrow R_{T2} \rightarrow R_{T4}
$$

Each text description is added to a visual constraints prompt $\{V_{T1}, V_{T2}, V_{T3}, V_{T4}\}$ and passed to the image generator $f_G$. $V_{T1}$ instructs $f_G$ to generate a photorealistic eye-level photograph of the base environment only, with no workers or equipment, a clean minimalist background, and no overlaid text or UI elements. $V_{T2}$ instructs $f_G$ to modify $I_{T1}$ via image-to-image generation, integrating workers and equipment seamlessly into the existing scene while locking the camera angle, perspective, and focal length to maintain visual consistency with $I_{T1}$. $V_{T2}$ also embeds $R_{T4}$ as a spatial positioning constraint, requiring that workers and equipment be placed such that the hazard described in $R_{T4}$ can realistically occur in the subsequent frame, without depicting the hazard itself. $V_{T3}$ instructs $f_G$ to modify $I_{T2}$ by adding two annotation layers only: a red outline or glow around the at-risk worker identified in $R_{T3}$, and a high-contrast safety warning banner displaying the warning phrase in the upper or lower third of the image. $V_{T4}$ instructs $f_G$ to modify $I_{T2}$ to depict the hazard event, allowing the camera angle to be adjusted so that the worker-hazard interaction is clearly visible while preserving the background scene. The complete text of all four stage prompts and the corresponding visual constraint prompts is available in the supplemental files sections~A.1 and~A.2. The full temporal image generation pipeline can therefore be represented as:
$$
I_{T1} = f_G\left(V_{T1} + R_{T1}\right)
$$
$$
I_{T2} = f_G\left(I_{T1},\ V_{T2} + R_{T2} + R_{T4}\right)
$$
$$
I_{T3} = f_G\left(I_{T2},\ V_{T3} + R_{T3}\right)
$$
$$
I_{T4} = f_G\left(I_{T2},\ V_{T4} + R_{T4}\right)
$$

\begin{figure}[H]
    \centering
    \includegraphics[width=0.9\textwidth]{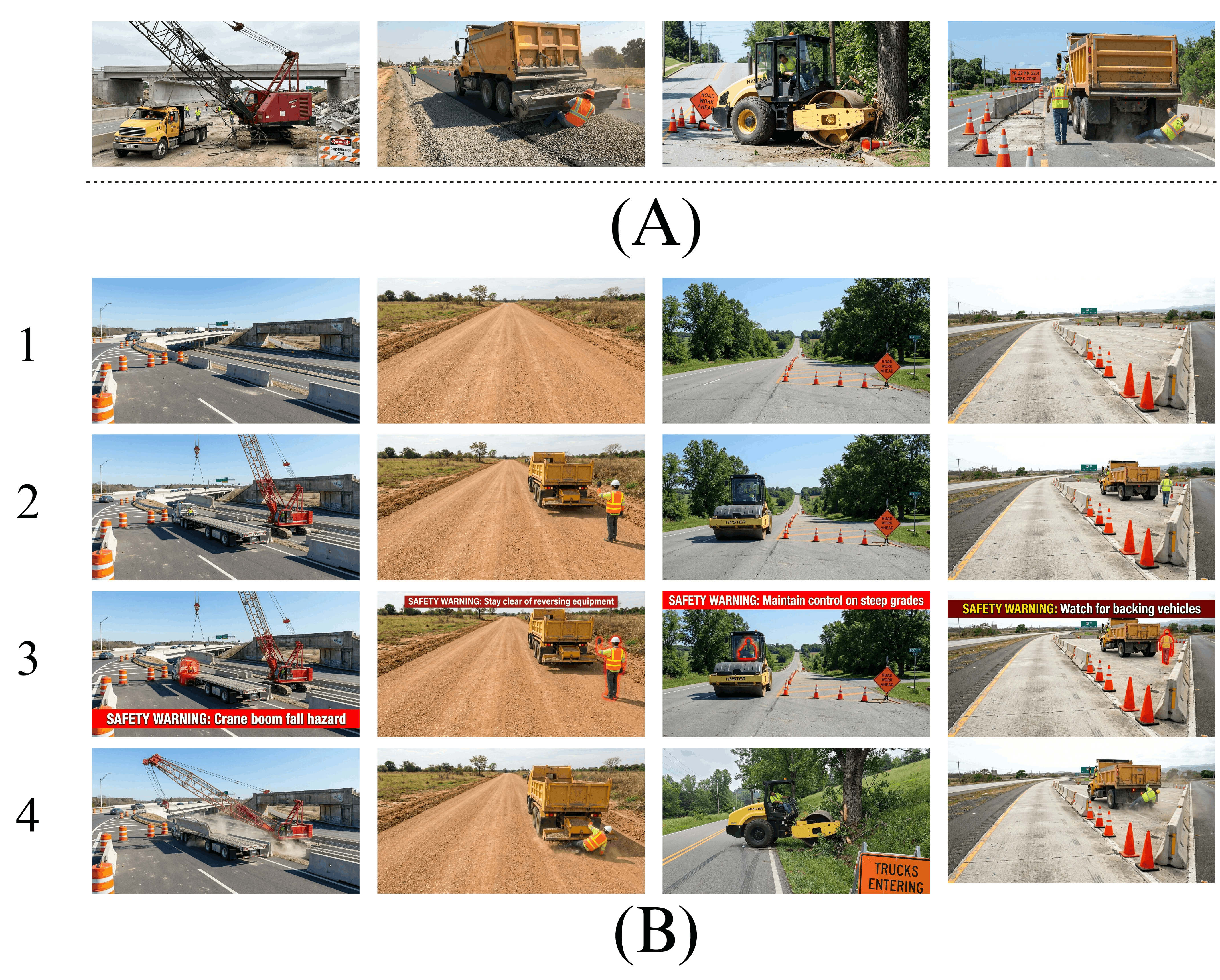}
    \caption{Examples of highway construction images generated from the SIR database. (A) single-pass images and (B) temporal sequences. Stages of temporal sequence indicated by numbers (1, 2, 3, and 4).}
    \label{fig:fig2}
\end{figure}

Figure~\ref{fig:fig2} (B) presents four representative temporal image sequences, each consisting of four frames generated from a single SIR incident record (corresponding to the same four incidents shown in panel A). Across all four sequences, the infrastructure and scene geometry established in the first frame are consistently preserved in subsequent rows, demonstrating the visual continuity maintained by the iterative image-to-image conditioning pipeline. A step-by-step worked example of the temporal pipeline, including all four verbatim descriptions and both iterations, is provided in the supplemental files section~C (Figure~A3). 

\subsection{Evaluation}
\label{sec:eval}
Two complementary evaluation strategies were employed to assess the quality of the synthetic highway construction safety images. The first was an automated evaluation using CLIP (Contrastive Language--Image Pre-training)~\cite{radford2021learning}, which measures the semantic alignment between generated images and their source text descriptions in a shared embedding space. The second was an expert-based evaluation, in which domain experts rated generated images across multiple quality dimensions. Together, these evaluations provide both quantitative retrieval metrics and qualitative assessments of image quality.

\subsubsection{CLIP-Based Semantic Alignment Evaluation}
To assess how well the generated images capture the semantic content of their source incident descriptions, a retrieval evaluation was performed using OpenAI's CLIP ViT-L/14 model~\cite{radford2021learning}, the most widely used image-text embedding baseline. The objective of this analysis was not to optimize retrieval performance, but to test whether matched image-description pairs are semantically distinguishable from mismatched pairs. CLIP was not fine-tuned for this task, so any retrieval performance above chance reflects semantic content recoverable by a general-purpose model rather than domain-specific knowledge. The CLIP model employed herein limits all text-based inputs to 77 tokens (approximately 50-60 words). Accordingly, using the entire SIR record as the text-embedding component was not feasible. The text label for this analysis concatenates the "event\_desc" field, which provides a concise event summary (e.g., "Employee's leg fractured when struck by falling beam"), with the abstract field, truncated to fit the 77-token window. Prior to concatenation, the abstract field was preprocessed with a series of regular expression substitutions to remove common patterns such as date/time preambles, employee identifiers, and company names, thereby allocating the remaining token budget to hazard-relevant content. Additionally, for temporal sequence–based images, only the stage 4 image ($I_{T4}$) was embedded for evaluation. The assumption was that any semantic errors present in $I_{T1}$ or $I_{T2}$ propagate through the image-to-image editing process, making evaluation of $I_{T4}$ alone an adequate proxy for assessing the entire sequence.

Cosine similarity in CLIP's shared 768-dimensional embedding space was computed between L2-normalized image and text embeddings:

$$
\text{sim}(I, T) = \mathbf{e}_I \cdot \mathbf{e}_T
= \sum_{d=1}^{768} e_{I,d} \, e_{T,d}
$$

\citet{liang2022mind} demonstrated that CLIP's image and text encoders embed their outputs into separate regions of the shared space, producing a positive baseline similarity even between unrelated images and descriptions. As a result, a cosine similarity between an image and a text must be compared against the baseline similarity of unrelated pairs to determine whether the match is meaningful.

Cosine similarity-based evaluation was measured in the text-to-image retrieval direction, in which each of the 75 preprocessed text descriptions served as a query and was directly compared against a gallery of 300 synthetic images. This gallery comprised, for each incident, two single-pass hazard images and two temporal stage 4 hazard images, yielding four correct-target images per query. As such, for each $D$, 300 images $I$ were ranked by cosine similarity, and retrieval performance was evaluated against these four correct targets. Retrieval was additionally evaluated separately by generation mode, in which each query ranked within that mode's 150-image gallery against its two correct targets.

An empirical null distribution was constructed from all mismatched (non-corresponding) image-text pairs. That is, the 75 text queries correspond to 296 non-corresponding images, yielding $75 \times 296 = 22{,}200$ mismatched pairs. Additionally, a distribution was constructed from all matched pairs, since each of the 75 text queries corresponds to 4 images, yielding $75 \times 4 = 300$. The gallery comprises one image per generated output: the two single-pass images and the final hazard frame ($I_{T4}$) for each of the two temporal sequences produced for a given record. Thus, the 300-image gallery is a subset of the 750 images generated overall. For each text query, all 300 images are ranked by cosine similarity; reciprocal rank is computed based on the best-ranked relevant image; and recall at $k$ (R@$k$) is the proportion of queries for which at least one relevant image appears in the top $k$. 

Cosine similarity between these embeddings in the shared CLIP space was characterized by its mean and standard deviation. A one-sided Mann-Whitney $U$ test was used to compare the distribution curve of matched and mismatched pairs. Because each image creates multiple mismatched pairs, the independence of observation assumption of this test is only partially satisfied. P-value from this test would indicate that the generated images encode semantic content beyond what a visually similar but contextually different construction safety scene would. The practical magnitude of that separation is quantified by Cohen's $d$:

$$
d = \frac{\mu_{\text{match}} - \mu_{\text{mismatch}}}
{\sqrt{\dfrac{(n_{\text{match}}-1)\sigma^2_{\text{match}} 
+ (n_{\text{mismatch}}-1)\sigma^2_{\text{mismatch}}}
{n_{\text{match}} + n_{\text{mismatch}} - 2}}}
$$
\noindent where $\mu$, $\sigma$, and $n$ denote the mean, standard deviation, and sample size of each group, respectively. The widely used convention is that $d = 0.2$ is considered a small effect, $d = 0.5$ a medium effect, and $d = 0.8$ a large effect~\cite{cohen2013statistical}. Additionally, retrieval performance was quantified using Mean Reciprocal Rank (MRR) and Recall at $k$ (R@$k$). MRR measures the average inverse rank of the first correct result, and R@$k$ measures the proportion of queries for which at least one correct item appears in the top $k$ results.

\subsubsection{Expert-Based Image Evaluation}
A panel of six domain experts with experience in highway construction safety, comprising three practicing construction professionals and three civil engineering students, independently evaluated the AI-generated image sets. The construction professionals held the job titles "Assistant Highway Maintenance Manager," "Safety Engineer", and "Civil Engineering Manager." All three have work experience with state Departments of Transportation or the Federal Highway Administration (U.S.A.). The student reviewers were two graduate students and one undergraduate student. All students had formal research experience regarding construction safety. Each image set was assessed by at least two reviewers. Due to varying availability among expert evaluators, the number of evaluations completed per rater was not uniform; however, the minimum dual-rater coverage across all items ensured that no assessment relied on a single rater's judgment. 

Reviewers were presented with a pair of images (or a pair of temporal sequences), both generated using the same image-generation methodology (i.e., single-pass or temporal) and both generated from the same SIR record ($D$). Generated images were evaluated using a structured assessment comprising three distinct sections. This assessment was administered through a web-based form; a sample screenshot of the evaluation interface is provided in the supplemental files section~B (Figure~A1). The first section was a checklist-based hallucination search. A hallucination, in this context, refers to a feature in the image that does not accurately match the input incident description or the construction work zones in general. Following a curated list of typical hallucinations relevant to highway construction safety hazard images, reviewers identified whether each hallucination type was present in the image. Providing a binary, checklist-based criterion for image evaluation is a more structured and effective approach for identifying errors than ad hoc identification~\cite{chaudhary2025prompt, cook2024ticking}. Images were assessed across four dimensions (single-pass) or six dimensions (temporal):
\begin{enumerate}
    \item Processing Artifacts: watermarks, overlaid text, unwanted photographer elements.
    \item Hazard Accuracy: depicted hazard matches the SIR record (Steps 2–4 for sequences).
    \item Scene Realism: logically arranged elements consistent with a real highway work zone job site.
    \item Visual Coherence: physically plausible renderings.
    \item Temporal Consistency*: stable people/equipment across Steps 2–4.
    \item Hazard Alert Accuracy*: Step 3 overlay correctly identifies and describes the at-risk worker.
\end{enumerate}
\noindent\footnotesize{*Temporal sequence evaluation only.}
\normalsize

\begin{figure}[H]
    \centering
    \includegraphics[width=0.9\textwidth]{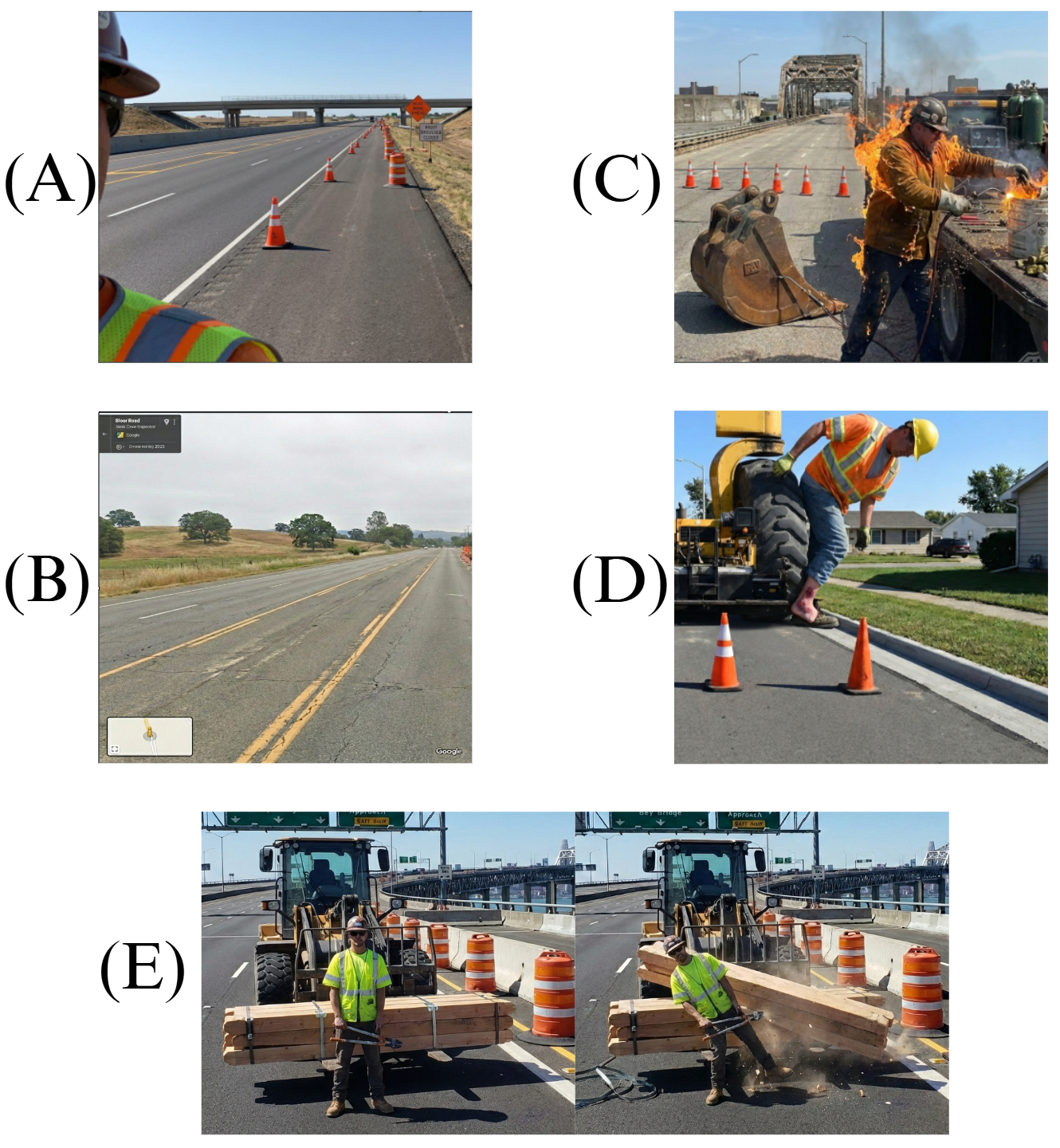}
    \caption{Examples of hallucinations by category: (A) processing artifact: the head and shoulder of a worker taking a photo are visible, (B) processing artifact: Google Street View watermarks are visible, (C) scene realism: work zone setup is illogical, and the welding tool is plugged into the excavator head, (D) visual coherence: worker and machine wheel are abnormally large, (E) temporal consistency: bundle of wood duplicates between images.}
    \label{fig:fig5}
\end{figure}

For all questions, a yes/no response was required, with "yes" indicating that the specific type of hallucination was not in the image/sequence. When a no response was given, the rater was instructed to explain the hallucination. Figure~\ref{fig:fig5} (A-E) shows several examples of the hallucinations described in the expert survey. 

Once the reviews were completed, for each binary checklist item (coded as 1 if the image was free of the specified hallucination or 0 if the hallucination was present), the pass rate was computed as the percent of all individual ratings coded as 1 across both iterations within each method (single-pass and temporal). To quantify inter-rater reliability, Fleiss's $\kappa$ was computed for each dimension. 
Fleiss's $\kappa$ was selected because reviewers were not crossed with items: most image sets were evaluated by two reviewers, while a subset received three or four evaluations due to varying reviewer availability, and the reviewers assigned to any given set were not fixed. Agreement strength was interpreted using the Landis and Koch thresholds: below 0.20 as slight, 0.21–0.40 as fair, 0.41–0.60 as moderate, 0.61–0.80 as substantial, and above 0.80 as almost perfect~\cite{landis1977measurement}. 

The second phase of the evaluation was a direct comparison of educational utility between the two images. Reviewers assessed the overall suitability of each image as an educational visualization of the described construction hazard, using a three-tier scale: "No Issues -- Fully Acceptable", "Minor Issues -- Still Acceptable", and "Major Issues -- Not Acceptable". Reviewers were also asked to provide written justification for their educational utility ratings. Educational ratings were defined as the image's ability to serve as a visual representation of the hazard in classroom discussions. Hallucinations in a synthetic image do not necessarily affect the educational value. Accordingly, reviewers were instructed that hallucinations did not automatically invalidate the educational value if the hazard interaction remained clear. To examine which hallucination types most strongly influenced educational utility judgments, checklist failure rates were cross-tabulated against the three educational utility categories. 

The third and final phase, following the best-practice recommendations of \citet{otani2023toward} for human evaluation of text-to-image generation, had reviewers provide scalar ratings of the hazard image (Step 4 for sequences; the single image for single-pass) for fidelity and alignment. Fidelity was rated on a 1-5 Likert scale, anchored by "Looks like an AI-generated photo" and "Looks like a real photo". Alignment (that is, semantic agreement with the SIR report) was rated on a 1-5 Likert scale anchored by "Does not match at all" and "Matches exactly". For both scales, means and standard deviations were computed across all ratings pooled within each methodology. To characterize inter-rater variability on the Likert scales, the mean absolute difference between paired reviews ($|\Delta|$) was reported, representing the average magnitude of disagreement in scale points between two reviewers evaluating the same image.

\section{Results}
After applying the image generation methodology to both Mode I (single-pass) and Mode II (temporal) to the 75 randomly selected OSHA narratives, 750 images were created in total. This comprises 150 single-pass images (each of the 75 narratives producing two images) and 600 temporal images (each of the 75 narratives was used twice to create two sequences of four images). All images were of resolution $1408\times768$, consuming an average disk space of 933 KB and a total disk space of 684 MB.

\subsection{CLIP-based Semantic Alignment}
\begin{figure}[H]
    \centering
    \includegraphics[width=0.9\textwidth]{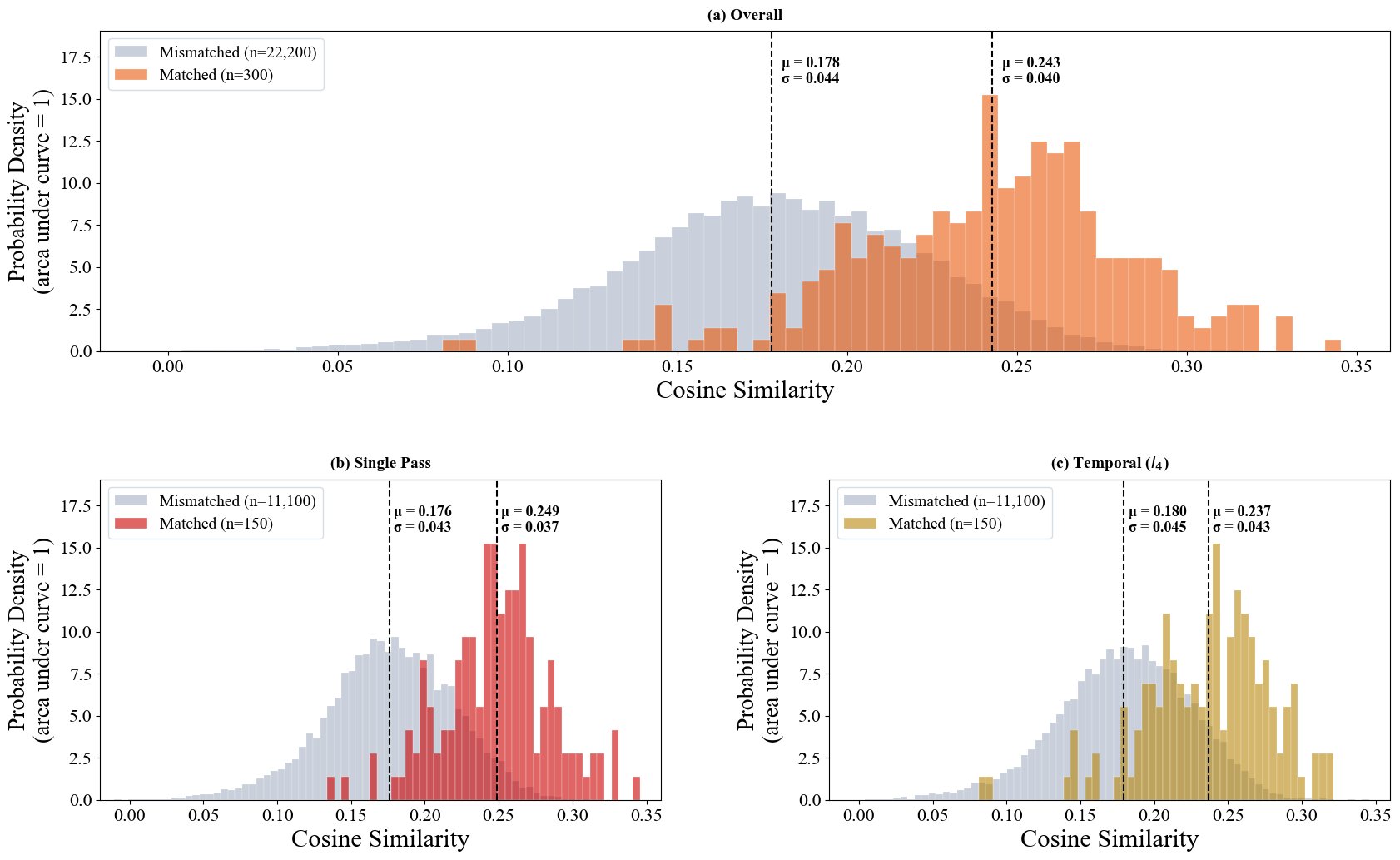}
    \caption{Distribution of cosine similarity scores between CLIP-encoded text descriptions and generated images: (A) overall, (B) single-pass images, and (C) temporal images, segmented into corresponding and non-corresponding pairs.}
    \label{fig:fig3}
\end{figure}

The empirical null distribution, constructed from 22,200 mismatched image-text pairs and visualized in Figure~\ref{fig:fig3} (A), was characterized by a mean cosine similarity of $\mu = 0.178$ and a standard deviation of $\sigma = 0.044$. The positive null mean is consistent with the modality gap phenomenon described by \citet{liang2022mind}, in which CLIP's image and text encoders embed their outputs onto geometrically separated cones in the shared space, producing a positive baseline similarity even between unrelated pairs. Matched image/text pairs had a mean cosine similarity of $\mu = 0.243$, and a standard deviation of $\sigma = 0.040$. The matched mean therefore sits $1.48$ null standard deviations above the null mean. This result indicates that the image generation model encoded the source incident description into the synthetic image. The one-sided Mann-Whitney $U$ test yielded a highly significant $p$-value ($1.22\times10^{-106}$). This conclusion is supported by a Cohen's d of 1.477, which is substantially greater than the baseline for a "large" effect size. Separated by single-pass and temporal (see Figure~\ref{fig:fig3}~(B) and (C)), the effect size remains large for both single-pass ($d=1.692$) and temporal ($d=1.274$), with the null hypothesis rejected in both cases ($p<10^{-43}$). Full distributional statistics for the overall and per-methodology comparisons are reported in the supplemental files (Table~A1).

The substantial overlap between matched and mismatched distributions in Figure~\ref{fig:fig3} reflects the homogeneity of the construction-safety domain: because all images and queries originate from the same domain, even non-corresponding pairs share substantial visual and textual content. The meaningful result, therefore, is the consistent rightward shift in matched pairs and the statistically significant effect size.

Table~\ref{tab:tab2} summarizes the text-to-image retrieval performance. The pipeline achieved an MRR of 0.602 for text/image matches overall, an R@1 of 46.7\%, and an R@10 of 84.0\%, meaning that over four-fifths of all queries had at least one of their relevant image ranked within the top 10 of 300 candidates. Comparing the two generation modes, each ranked within its own 150-image gallery against its two relevant images, the single-pass variant outperformed the temporal variant on every metric, with an MRR of 0.612 versus 0.455 and R@1 of 49.3\% versus 30.7\%. This pattern held across all recall depths, with the single-pass variant achieving 86.7\% at R@10, compared to 73.3\% for the temporal variant. The advantage in MRR, the primary retrieval metric, was statistically significant (difference of 0.157, 95\% CI [0.065, 0.249]; query-level bootstrap p = 0.0012). The single-pass advantage was also significant at R@1 (difference of 18.7\%, $p = 0.0032$), R@5 (difference of 18.7\%, $p = 0.0010$), and R@10 (difference of 13.3\%, $p = 0.0222$).

\begin{table}[H]
    \centering
    \footnotesize
    \setlength{\tabcolsep}{5pt}
    \caption{Text-to-image retrieval performance.}
    \label{tab:tab2}
    \begin{tabular}{l ccc cccc}
        \toprule
        \textbf{Group}
        & \textbf{Queries}
        & \textbf{Candi-}
        & \textbf{Rel./}
        & \textbf{MRR}
        & \textbf{R@1}
        & \textbf{R@5}
        & \textbf{R@10} \\
        & & \textbf{dates} & \textbf{query} & & \textbf{(\%)} & \textbf{(\%)} & \textbf{(\%)} \\
        \midrule
        Overall & 75 & 300 & 4
            & 0.602 & 46.7 & 74.7 & 84.0 \\
        & & & & {\scriptsize [0.512, 0.689]} & {\scriptsize [34.7, 57.3]}
              & {\scriptsize [64.0, 84.0]} & {\scriptsize [74.7, 92.0]} \\
        \addlinespace
        Single-pass & 75 & 150 & 2
            & 0.612 & 49.3 & 78.7 & 86.7 \\
        & & & & {\scriptsize [0.518, 0.702]} & {\scriptsize [37.3, 60.0]}
              & {\scriptsize [69.3, 88.0]} & {\scriptsize [78.7, 93.3]} \\
        \addlinespace
        Temporal & 75 & 150 & 2
            & 0.455 & 30.7 & 60.0 & 73.3 \\
        & & & & {\scriptsize [0.368, 0.544]} & {\scriptsize [21.3, 41.3]}
              & {\scriptsize [49.3, 70.7]} & {\scriptsize [62.7, 82.7]} \\
        \bottomrule
    \end{tabular}
\end{table}

\subsection{Expert-based Evaluation}
Table~\ref{tab:expert-summary}~(A) summarizes the checklist-based hallucination assessment, with inter-rater agreement statistics reported in the supplemental files (Table~A2(A)). Processing artifacts were the least prevalent hallucination type, with pass rates of 94.9\% for single-pass images and 84.1\% for temporal sequences, and inter-rater agreement in the moderate-to-substantial range. Hazard accuracy, which assessed whether the depicted hazard interaction reflected the incident described in the source SIR record, was the weakest dimension for both modes, with pass rates of 62.3\% and 60.6\% for single-pass and temporal images, respectively. Inter-rater agreement on hazard accuracy was fair for single-pass and moderate for temporal, with Fleiss's $\kappa$ values of 0.343 and 0.411, suggesting that reviewers applied moderately different thresholds when judging semantic alignment. Scene realism exhibited the lowest inter-rater agreement among all dimensions ($\kappa < 0.23$ in both modes), despite a 79.3\% pass rate for single-pass images. For temporal sequences, scene realism dropped to 54.7\%, as expected, since generating more images increases the likelihood that at least one will contain an unrealistic feature. Temporal consistency, assessed only for temporal sequences, had a pass rate of 60.3\% with moderate agreement ($\kappa = 0.494$), the highest among the subjective dimensions. Hazard alert accuracy, which evaluated whether the Step 3 warning overlay correctly identified the at-risk worker and described the impending hazard, achieved the highest temporal-specific pass rate at 86.5\% and moderate inter-rater agreement.

\begin{table}[H]
    \centering
    \small
    \caption{Expert evaluation summary by methodology.
    (A) Per-dimension pass rates and fail-in-major rates from the checklist assessment. (B) Overall educational acceptability and Likert-scale fidelity and alignment ratings.}
    \label{tab:expert-summary}

    \textbf{(A)}\\[4pt]
    \begin{tabular}{l cc cc}
        \toprule
        & \multicolumn{2}{c}{\textbf{Pass Rate (\%)}}
        & \multicolumn{2}{c}{\textbf{Fail-in-Major (\%)}} \\
        \cmidrule(lr){2-3} \cmidrule(lr){4-5}
        \textbf{Dimension}
        & \textbf{Single-Pass} & \textbf{Temporal}
        & \textbf{Single-Pass} & \textbf{Temporal} \\
        \midrule
        Processing Artifacts  & 94.9 & 84.1 & 11.1 & 19.7 \\
        Hazard Accuracy       & 62.3 & 60.6 & 93.7 & 75.8 \\
        Scene Realism         & 79.3 & 54.7 & 34.9 & 51.5 \\
        Visual Coherence      & 87.7 & 82.1 & 15.9 & 28.0 \\
        Temporal Consistency  & ---  & 60.3 & ---  & 61.4 \\
        Hazard Alert Accuracy & ---  & 86.5 & ---  & 25.8 \\
        \bottomrule
    \end{tabular}

    \vspace{10pt}

    \textbf{(B)}\\[4pt]
    \begin{tabular}{l cc}
        \toprule
        \textbf{Measure}
        & \textbf{Single-Pass}
        & \textbf{Temporal} \\
        \midrule
        Fully Acceptable (No Issues)         & 47.0\% [39.5, 54.8] & 27.4\% [21.7, 33.4] \\
        Minor Issues (Still Acceptable)      & 34.1\% [27.5, 40.7] & 33.6\% [28.1, 39.2] \\
        Major Issues (Not Acceptable)        & 18.9\% [13.6, 24.5] & 38.9\% [31.7, 46.2] \\
        Overall Acceptability                & 81.1\% [75.4, 86.5] & 61.0\% [53.8, 68.2] \\
        \midrule
        Fidelity (Likert, $\mu \pm \sigma$)  & $4.14 \pm 1.05$     & $3.51 \pm 1.35$ \\
        Alignment (Likert, $\mu \pm \sigma$) & $4.07 \pm 1.13$     & $3.94 \pm 1.24$ \\
        \bottomrule
    \end{tabular}
\end{table}
Table~\ref{tab:expert-summary}~(B) reports overall educational acceptability by methodology. Single-pass images were rated Fully Acceptable in 47.0\% of evaluations and Acceptable with Minor Issues in 34.1\%, yielding an overall acceptance rate of 81.1\%. Temporal sequences scored 27.4\% Fully Acceptable and 33.6\% Minor Issues, for an overall acceptance rate of 61.0\% (full distribution in the supplemental files, Table~A2(B)). The 20.0 percentage-point advantage for single-pass images was statistically significant (95\% CI [$+11.1$, $+29.1$], $p<0.0001$). The Fail-in-Major column of Table~\ref{tab:expert-summary}~(A) reports, for each dimension, the proportion of "Major Issues" ratings in which that dimension was flagged, revealing which hallucination types most strongly correlate with educational rejection. Hazard accuracy emerged as the dominant predictor of educational rejection: for single-pass images rated "Major Issues," 93.7\% had a hazard accuracy failure, compared to only 0.6\% of images rated "Fully Acceptable." For temporal sequences, the same pattern held at 75.8\% versus 4.3\%. Scene realism and visual coherence failures, by contrast, plateaued between the "Minor Issues" and "Major Issues" categories for single-pass images, suggesting that reviewers sometimes tolerate these issues when the hazard itself is correctly depicted. For temporal sequences, temporal consistency was a strong driver of rejection, with a failure rate of 61.4\% among "Major Issues" ratings compared to 8.6\% among "Fully Acceptable" ratings.

Likert-scale fidelity and alignment ratings are reported in Table~\ref{tab:expert-summary}~(B). Single-pass images received a mean fidelity rating of 4.14 (SD = 1.05). For alignment, single-pass images scored a mean of 4.07 (SD = 1.13). The mean absolute differences were 0.84 and 0.89 for fidelity and alignment, respectively, indicating that paired reviewers typically differed by approximately 1 scale point, consistent with the inherent subjectivity of holistic image quality assessment. For temporal image sequences, the mean responses for fidelity and alignment were 3.51 and 3.94, with standard deviations of 1.35 and 1.24, respectively. As with single-pass images, the mean absolute difference was approximately 1 scale point. 

The key difference in results between modes is that the lower scores seen from temporal images were most highly concentrated in fidelity rather than alignment. Average fidelity fell by 0.63 points (95\% CI [$+0.38$, $+0.89$], $p<0.0001$), whereas the corresponding average reduction in alignment was 0.13 points and was not statistically significant (95\% CI [$-0.15$, $+0.39$], $p = 0.379$). Generating synthetic images as a sequence, accordingly, increases the odds that the output scores lower on photorealistic quality, but does not proportionally reduce the semantic alignment between the generated image and its source incident description. This suggests that the image-to-image conditioning pipeline introduces visual artifacts and degradation that reviewers perceive as less realistic, while largely preserving the scene's hazard-relevant content.

A stochastic recovery analysis was conducted based on the comparative evaluations (see the supplemental files, Table~A2(C)). For each paired assessment, the two iterations were categorized independently by each evaluator. As such, a pair of images (or sequences) can be categorized as "Both acceptable", "Mixed", or "Both rejected". For single-pass images, 70.1\% of paired assessments had both iterations rated as acceptable, 22.2\% were mixed, and only 7.8\% had both iterations rejected. For temporal sequences, 44.4\% had both iterations acceptable, 33.1\% were mixed, and 22.5\% had both rejected. The recovery rate, defined as the proportion of mixed pairs among all pairs in which at least one iteration was rejected, was 74.0\% for single-pass and 59.6\% for temporal. 

\section{Discussion}
The evaluation results indicated that generative AI has strong capabilities to create useful safety hazard visualizations that do not currently exist and can therefore improve current health and safety training curricula. The first principal finding of this work was that CLIP-based analysis confirmed that the synthetic images are encoded semantically with respect to the safety content beyond chance (Cohen's $d = 1.477$). Despite the visual homogeneity of the dataset, matched image-text pairings were statistically distinguishable from mismatched pairs across all conditions ($p<10^{-43}$; see Figure~\ref{fig:fig3}), thereby validating the methodology for generating semantically retrievable images. Retrieval performance also exceeded chance by a wide margin. With four relevant images among 300 candidates, a random ranking would return a relevant image at the first position for 1.3\% of queries and within the top ten for 12.7\%; the pipeline achieved 46.7\% and 84.0\% respectively, and the best-ranked relevant image typically appeared within the first two results (MRR $= 0.602$).It should be noted that CLIP was not tuned on construction safety imagery, and its 77-token input limit required truncation of the incident narratives, both of which likely suppressed retrieval performance. 

A second principal finding was that the domain experts rated a majority of the generated images as educationally acceptable for use in safety training. Single-pass images achieved an overall acceptability rate of 81.1\%, compared with temporal images, which had a lower but still majority acceptability rate of 61.0\%. Cross-tabulating checklist failures against educational utility ratings revealed that hazard accuracy was the primary determinant of whether an image was judged educationally viable. Hazard accuracy failure rates increased substantially from the "Fully Acceptable" to the "Major Issues" categories in both modes, whereas scene realism and visual coherence failures did not follow the same trend. This suggests that reviewers tolerate imperfections in the surrounding environment when the hazard interaction itself is correctly depicted. Additionally, stochastic recovery analysis revealed that when at least one iteration was rejected, the alternative iteration generated from the same record and identical prompt was acceptable in 74.0\% of single-pass cases and 59.6\% of temporal cases. This indicates that generation quality is determined not only by the source incident record but also by stochasticity in image generation. As such, future research on developing an automated system to identify and regenerate unacceptable images would be a valuable contribution to this domain. 

Thirdly, single-pass images outperform temporal sequences across all quantitative metrics. Temporal sequences offer a unique capability in construction education: narrative progression and storytelling, which can increase engagement and knowledge retention~\cite{eggerth2018evaluation}. However, because each frame in the temporal pipeline is conditioned on the output of the previous step, errors introduced in earlier stages propagate throughout and compound, increasing the likelihood that at least one frame in the sequence contains a hallucination. For instance, Street View interface artifacts that occasionally appeared in Stage 1 outputs persisted across all subsequent frames. Temporal sequences were also subject to two additional checklist dimensions (temporal consistency and hazard alert accuracy) and thus to additional scrutiny. Notably, this compounding degradation manifests primarily in fidelity rather than alignment. The findings in Table~\ref{tab:expert-summary} indicate an additional source of compounding degradation: temporal consistency was the second-greatest driver of educational rejection. 

Beyond the image-generation methodology itself, this study contributes an expert-based evaluation framework for synthetic construction-safety imagery. The checklist-based hallucination assessment, informed by \citet{cook2024ticking}'s finding that structured criteria improve evaluation consistency, decomposed image quality into inspectable dimensions rather than holistic judgments. Inter-rater agreement, as measured by Fleiss's kappa, varied substantially across dimensions (see the supplemental files, Table~A2(A)). Dimensions involving concrete judgments, such as processing artifacts and temporal consistency, achieved moderate to substantial agreement, whereas more subjective dimensions, such as scene realism and visual coherence, exhibited only slight to fair agreement. This pattern suggests that the latter dimensions, as currently defined, may be too broad to elicit consistent judgment between experts.

These results support a workflow for incorporating synthetic images into existing safety-training procedures. First, select an incident narrative relevant to the session's training goals for the session from either the SIR database or an internal log. Second, generate two iterations of both temporal and single-pass images based on the incident narrative. Given that single-pass image pairs had at least one acceptable image 92.3\% of the time (See Table~A2(C)), generating two iterations per incident record will, in most cases, yield at least one educational, viable image without prompt refinement. Third, a member of the training team should review the output using the presented checklist and determine which images are educationally acceptable. The lower acceptability of temporal sequences (61.0\%) implies that it may take more iterations to find an effective image. If none of the generated images are sufficient, the SIR report or the prompt can be adjusted, with specific targets aimed at the errors identified in the expert screening. Once useful images are created, they can be integrated into the toolbox talk, pre-task briefing, or safety presentation. By generating new synthetic images for the safety talk, the image-based and storytelling-based training format identified by \citet{burke2011dread} and \citet{cullen2008tell} as most effective for high-hazard occupations is leveraged.

Several relevant limitations must be acknowledged. First, the Mann-Whitney $U$ test assumes independence of observations, which is partially violated because each image appears multiple times in mismatched pairs. The reported $p$-value should be considered within that context, and the separation between matched and mismatched distributions is more appropriately judged by the effect size ($d = 1.477$), which does not depend on the independence of pairs and is considered a "large" effect size. Second, the results are specific to Google's Gemini 3 Pro image generation model as accessed in February 2026; given the rapid pace of development in frontier image generation, future models may address many of the hallucination types identified herein. Only one multimodal embedding model was evaluated for retrieval metrics (OpenAI's CLIP ViT-L/14), and performance may vary with newer models of different architectures. Additionally, the rater panel of six evaluators (three construction professionals and three civil engineering students) exhibited a substantial imbalance in rating volume: two reviewers completed approximately 84\% of all evaluations, while the remaining four contributed 16\%. This imbalance was a consequence of varying availability rather than a design choice; thus, this article cannot determine the extent to which these acceptability judgments generalize to the broader construction safety industry. Although all members of the rating panel held qualifying professional or research-based roles, the number of years of experience was not recorded and thus could not be reported in this study. The SIRs used herein comprise only 75 records; thus, this study is limited in its coverage of niche and underreported highway hazards. The evaluation assessed educational suitability, as judged by domain experts, and retrieval performance using CLIP-based analysis, but did not measure downstream training effectiveness. This study did not evaluate how the content- and construction-specific terminology in the safety reports affects image-generation outputs. Lastly, the generalizability to other construction sectors was not addressed, so it is unclear how closely the results would align with other industries.

These technical limitations also raise ethical considerations, since synthetic hazard imagery is designed to look realistic to teach workers what to look for; a misleading or inaccurate visual could instill incorrect knowledge or poor mental models of how hazards exist in the real world. Additionally, biases or visualization guardrails in AI models may affect the design of synthetic images; thus, if a teaching curriculum became too heavily reliant on them, blind spots could emerge. Although expert review partially mitigates inaccuracies or misrepresentations in hazard visualizations, the imagery should be reviewed holistically for representativeness and accuracy before deployment in a training course. To that end, synthetic imagery should supplement training courses rather than replace real incident documentation and expert instruction. 

Several directions for future research are enabled by this study. First, iterative prompt refinement using a secondary vision-language model to evaluate generated images and adjust the input prompt accordingly, as proposed by \citet{jeon2025iterative}, could improve educational acceptability. Second, a training effectiveness study comparing text-based, image-based, and synthetic-image-based instruction would help determine whether educational acceptability, as evaluated by domain experts, translates into knowledge retention. Third, the evaluation framework developed herein could be refined by decomposing scene realism and visual coherence into more specific sub-criteria to improve inter-rater agreement and by expanding the rater panel to include a broader range of construction safety professionals. Fourth, extending the temporal pipeline from image sequences to short generated videos would allow hazards to be depicted as continuous motion rather than as stages, though this would require addressing the same compounding-consistency challenges identified here at a higher frame rate. Lastly, expanding the methodology beyond NAICS code 237310 (Highway, Street, and Bridge Construction) or to other datasets in different scientific fields would test the generalizability of the proposed frameworks. 

\section{Conclusion}
This study addressed the research question of how generative AI can be used to create realistic, semantically valid imagery of highway construction safety hazards that evolve over time, and which modes should guide the evaluation of these images. Three contributions were presented. 

First, a structured prompt engineering framework for the single-pass generation of synthetic images depicting highway construction hazards was defined. Single-pass images achieved an overall educational acceptability rate of 81.1\% from the domain expert panel, with mean fidelity and alignment scores of 4.14/5 and 4.07/5, respectively. Additionally, a CLIP-based retrieval analysis showed that single-pass synthetic images exhibit statistically significant semantic alignment relative to the null mean. 

Second, a temporal image generation pipeline to produce four images in a sequence to depict (1) a safe base environment, (2) active construction, (3) early warning for a hazard event, and (4) a hazardous construction event leading to severe injury was developed. Temporal sequences achieved an educational acceptability rate of 61.0\%, with mean fidelity and alignment scores of 3.51/5 and 3.94/5, respectively. Similar to single-pass images, CLIP-based analysis revealed a statistically significant semantic alignment relative to the null mean. The results indicate that narrative-based synthetic visualization is feasible, but compounding errors across sequential images lead to more hallucinations than single-pass generation. 

Third, a multi-dimensional evaluation framework that combines CLIP-based semantic retrieval with an expert-based educational utility grade was developed. Cross-tabulation revealed that hazard accuracy was the dominant predictor of educational rejection across both modes. This finding provides clear guidance for future prompt engineering efforts: accurate depiction of the worker/hazard interaction matters more to educational value than photorealistic rendering of the surrounding scene.

These results demonstrate that generative AI can transform text-based narratives into visual training assets and address the data scarcity of images depicting dangerous construction activities. The methodologies enable safety instructors to produce project-specific, hazard-focused visualizations without photographing active hazards in real construction sites. Because AI-generated images are stochastic, practical deployment is not dependant on obtaining an acceptable image on the first attempt. Generating two iterations per incident record yielded at least one acceptable single-pass image in 92.3\% of cases. This makes the pipeline deployable by safety personnel using an existing SIR narrative or an internal incident log, with the checklist serving as a means to verify quality before integrating the image into a toolbox talk or safety briefing. As frontier image generation models continue to improve, the image generation framework and the evaluation guidelines established in this study provide a foundation for tracking how these improvements translate into gains in educational utility for highway construction safety training.

\subsection*{Funding}
The findings reported here are based on work performed with the support of the Impactful Resilient Infrastructure Science \& Engineering (IRISE) consortium in the Department of Civil and Environmental Engineering, Swanson School of Engineering at the University of Pittsburgh. We are indebted for the advice and assistance provided by the following representatives of IRISE member organizations that comprised the technical panel who guided work on the project: Pennsylvania Turnpike Commission, Pennsylvania Department of Transportation, Contractors Association of Western Pennsylvania, Michael Baker International, Allegheny County, Golden Triangle Construction, and the Federal Highway Administration.

\bibliography{bib}
\newpage
\begin{center}
  {\LARGE Generative AI for Visualizing Highway Construction Hazards Through Synthetic Images and Temporal Sequences (Supplemental Files)\par}
\end{center}

\appendix
\renewcommand{\thefigure}{A\arabic{figure}}
\renewcommand{\thetable}{A\arabic{table}}
\setcounter{figure}{0}
\setcounter{table}{0}
\section{Prompt Templates}
\label{appendix:prompts}
 
This section presents the complete set of prompt templates used in the image generation pipeline. Prompts are organized by generation stage. Placeholders enclosed in curly braces (e.g., \texttt{\{json\_data\}}) denote runtime substitutions drawn from the OSHA SIR record $D$ or from the output of a prior generation step. 
 
\subsection{LLM Text Generation Prompts ($f_\theta$)}
\label{appendix:text_prompts}
 
The following prompts are passed to the language model $f_\theta$ (Gemini 2.5 Flash) to generate the scene descriptions $\{R_{T1}, R_{T2}, R_{T3}, R_{T4}\}$ and $R_{SP}$.
 
\subsubsection*{Stage 1: Base Infrastructure ($P_{T1}$)}
 
\begin{quote}
\ttfamily\small
You are a Prompt Engineer for an AI image generator. \\
Describe the base infrastructure from this OSHA data: \{json\_data\}. \\\\
Instruction: Write 3 sentences focusing on the roadway, terrain, or traffic control layout (barriers, cones, lane markings). \\
Pre-construction phase: There should be no construction equipment or workers described yet. \\
Based on the Address and city, consider if this is an urban, highway, rural, or residential setting. \\
Constraint: Describe only the static environment. DO NOT include construction machines, vehicles, or workers. \\\\
Output Format: Plain text only. No preamble or markdown.
\end{quote}
 
\subsubsection*{Stage 2: Construction Activity ($P_{T2}$)}
 
\begin{quote}
\ttfamily\small
You are a Prompt Engineer for an AI image generator. \\
Add equipment and workers to the existing environment: \{infrastructure \_description\}. \\\\
Instruction: Write 3 sentences introducing the construction machines/ vehicles and 1--2 workers performing the routine task from the OSHA data (\{json\_data\}). \\
Focus: Mention equipment placement and worker positions relative to the infrastructure. Include PPE explicitly listed. \\
Consider the health and safety hazard that will be introduced in the next step, but do not describe it yet. What type of construction activity would occur before this hazard takes place? \\
Ensure that a worker, who will later interact with a hazard, is clearly depicted within the work zone. \\
Ensure that any construction machines which will later be involved in the hazard are also present and correctly positioned. \\
Tone: Objective and routine. Do not describe any danger yet. \\\\
Output Format: Plain text only. No preamble.
\end{quote}
 
\subsubsection*{Stage 3: Safety Warning Specification ($P_{T3}$)}
 
\begin{quote}
\ttfamily\small
You are a Prompt Engineer for an AI image generator. \\
Add a safety warning overlay to the existing scene: \{activity\_description\}. \\\\
Instruction: Write 2--3 sentences that: \\
1. Identify which worker is in the danger zone and should be highlighted with a red outline or glow. \\
2. Generate a concise safety warning phrase (5--7 words max) based on the upcoming hazard from the OSHA data: \{json\_data\}. \\\\
Safety Warning Examples: \\
- ``Watch for pinch points'' \\
- ``Watch for tipping vehicles'' \\
- ``Maintain safe distance from equipment'' \\
- ``Watch for overhead hazards'' \\
- ``Stay clear of swing radius'' \\
- ``Watch for struck-by hazards'' \\
- ``Beware of fall hazards'' \\\\
Output Format: \\
Line 1: Description of which worker to highlight and their position. \\
Line 2: SAFETY\_WARNING: [Your generated warning phrase] \\
Plain text only. No preamble.
\end{quote}
 
\subsubsection*{Stage 4: Hazard Event ($P_{T4}$)}
 
\begin{quote}
\ttfamily\small
You are a Prompt Engineer for an AI image generator. \\
Add a hazard description to the existing scene: \{activity\_description\}. \\
Describe the peak of the hazard and physical interaction based on this data: \{json\_data\}. \\\\
Instruction: Write 3 sentences capturing the moment the hazard becomes visible and the resulting physical interaction (e.g., impact or fall) between the worker and the threat. \\
Focus: Use the `event\_keyword' to define the specific threat and describe the mechanics of the event with forensic clarity. \\
Constraint: Do not dramatize. Stick to the physical description provided in the text. \\\\
Output Format: Plain text only. No preamble.
\end{quote}
 
\subsubsection*{Single-Pass Scene Description ($P_{SP}$)}
 
\begin{quote}
\ttfamily\small
You are a Prompt Engineer for an AI image generator. \\
Generate a forensic, frozen-moment description of a safety incident: \{json\_data\}. \\\\
Structure your description into three distinct sections, leading with the specific labels below: \\
1. THE INFRASTRUCTURE: Describe the base roadway or work zone layout (e.g., pavement, lane markings, traffic control). No machines or workers. \\
2. THE ACTIVITY: Introduce the construction equipment and workers performing their routine task within that space. \\
3. THE HAZARD: Describe the peak of the event, focusing on the physical interaction between the worker and the hazard (event\_keyword). \\\\
Constraints: \\
- 1--2 sentences per step. \\
- Start the response immediately with the Step 1 description. \\
- No dramatization or inventions. \\
- Output as a single cohesive paragraph. \\
- Plain text only. No preamble.
\end{quote}
 
\subsection{Image Generation Prompts ($f_G$)}
\label{appendix:image_prompts}
 
The following visual constraints prompts $\{V_{T1}, V_{T2}, V_{T3}, V_{T4}, V_{SP}\}$ are passed to the image generator $f_G$ (Gemini 3 Pro) alongside the scene descriptions from Section~\ref{appendix:text_prompts}. The placeholder \texttt{\{state\_description\}} denotes the corresponding scene description $R$ generated in the prior step.
 
\subsubsection*{Stage 1: Base Environment Image ($V_{T1}$)}
 
\begin{quote}
\ttfamily\small
You are generating an image taken at eye-level, as if an inspector took the photo while standing a few feet away. The worker taking the photo should be standing within the traffic control or work zone. \\
Generate a photorealistic educational visualization of a construction environment. \\
This state acts as the base environment; do not add any construction equipment or workers. \\
Only generate roadway, terrain, or traffic control layout (barriers, cones, lane markings) as specified in the description. \\
The construction zone must have open space for work to occur. \\
Description: \{state\_description\} \\\\
Crucial Constraints: \\
1. Style and Purpose: Use realistic photography with bright, even lighting suitable for an educational manual. \\
2. Minimize Clutter: The background should be relatively clean and minimalist. Only include equipment or structures explicitly mentioned in the description. Avoid random debris or complex textures that distract from the main subject area. \\
3. Text Rules: Do not include overlayed captions, watermarks, or UI elements. Text is only permitted if it appears naturally on environmental objects, such as safety signs, labels on equipment, or vehicle branding. \\
4. Do not have the GUI of Google Streetview overlayed on the image.
\end{quote}
 
\subsubsection*{Stage 2: Construction Activity Image ($V_{T2}$)}
 
\begin{quote}
\ttfamily\small
You are generating an image. \\
Acting as a VFX editor for an educational series, modify the provided input image to integrate the next step of the safety sequence. \\
This state introduces construction equipment and workers performing their routine task. \\
Modification Task (The Activity): \{state\_description\} \\\\
CRITICAL POSITIONING REQUIREMENT: \\
The following hazard will occur in the next state: \{hazard\_description\} \\
You MUST position the workers and equipment such that this hazard can realistically occur in the next image. \\
- Place the worker who will be involved in the hazard in a location and posture that makes the incident plausible. \\
- Position any equipment, vehicles, or objects involved in the hazard so they are oriented correctly for the event. \\
- Ensure spatial relationships (distances, angles, sight lines) support the described hazard mechanics. \\\\
All added construction activity must be within the work zone. \\\\
Crucial Constraints for Consistency and Clarity: \\
1. Lock Camera: Maintain the exact camera angle, perspective, and lens focal length of the input image. Do not zoom, pan, or reframe the shot. \\
2. Focal Point: The new elements (workers and construction equipment) must be the sharpest and most distinct part of the image. The existing background should remain distinct but secondary to the action. \\
3. Clean Integration: Add the new elements seamlessly into the existing geometry without adding unnecessary environmental clutter around them. \\
4. Text Rules: No overlayed text, labels, or captions. Text must only exist on physical objects within the scene, such as hazard signs or worker PPE. \\
5. Although the critical positioning requirement describes the hazard, this image must not depict the hazard event itself.
\end{quote}
 
\subsubsection*{Stage 3: Safety Warning Overlay Image ($V_{T3}$)}
 
\begin{quote}
\ttfamily\small
You are generating an image. \\
Acting as a VFX editor for an educational safety training series, modify the provided input image to add a safety warning overlay. \\
This state highlights the worker in danger and displays a safety notice before the hazard occurs. \\\\
Worker Highlight Description: \{state\_description\} \\\\
Crucial Constraints for Consistency and Clarity: \\
1. Lock Camera: Maintain the exact camera angle, perspective, and lens focal length of the input image. Do not zoom, pan, or reframe the shot. \\
2. Worker Highlight: Add a distinct red outline, glow, or semi -transparent red overlay around the worker identified as being in danger. The highlight should be clearly visible but not obscure the worker's details. \\
3. Safety Warning Banner: Display the safety warning text prominently in the upper or lower third of the image. Use a high-contrast format: white or yellow bold text on a red or dark background banner. The text should be large and readable. The safety warning text should start with ``SAFETY WARNING''. \\
4. Preserve Scene: Do not alter the existing workers, equipment, or environment. Only add the highlight effect and text overlay. \\
5. Educational Tone: The overall effect should resemble a training video freeze-frame or safety manual illustration with clear visual callouts.
\end{quote}
 
\subsubsection*{Stage 4: Hazard Event Image ($V_{T4}$)}
 
\begin{quote}
\ttfamily\small
You are generating an image. \\
Acting as a VFX editor for an educational series, modify the provided input image to integrate the next step of the safety sequence. \\
This state introduces the health and safety hazard and the worker's interaction with it. \\
Modification Task (The Hazard): \{state\_description\} \\
Do not create any new workers; the hazard must involve an existing worker from the previous state. \\
Do not create new construction equipment; the hazard must involve existing equipment from the previous state. \\\\
Crucial Constraints for Consistency and Clarity: \\
1. Move Camera: Feel free to adjust the camera angle so that the health and safety hazard and worker interaction are clearly visible. Maintain a realistic eye-level perspective. Do not change background info. \\
2. Focal Point: The new elements (the hazard or the incident) must be the sharpest and most distinct part of the image. The existing background should remain distinct but secondary to the action. \\
3. Clean Integration: Add the new elements seamlessly into the existing geometry without adding unnecessary environmental clutter around them. \\
4. Text Rules: No overlayed text, labels, or captions. Text should only exist on physical objects within the scene, such as hazard signs or worker PPE.
\end{quote}
 
\subsubsection*{Single-Pass Hazard Image ($V_{SP}$)}
 
\begin{quote}
\ttfamily\small
You are generating an image taken at eye-level, as if an inspector took the photo while standing a few feet away. \\
Generate a photorealistic educational visualization of a construction scenario. \\\\
Image Description: \{state\_description\} \\\\
Crucial Constraints for Consistency and Clarity: \\
1. Lock Camera: Maintain the exact camera angle, perspective, and lens focal length of the input image. Do not zoom, pan, or reframe the shot. \\
2. Focal Point: The new elements (workers, the hazard, or the incident) must be the sharpest and most distinct part of the image. The existing background should remain distinct but secondary to the action. \\
3. Clean Integration: Add the new elements seamlessly into the existing geometry without adding unnecessary environmental clutter around them. \\
4. Text Rules: No overlayed text, labels, or captions. Text should only exist on physical objects within the scene, such as hazard signs or worker PPE.
\end{quote}

\section{Evaluation Interface}
\begin{figure}[H]
    \centering
    \includegraphics[width=0.7\textwidth]{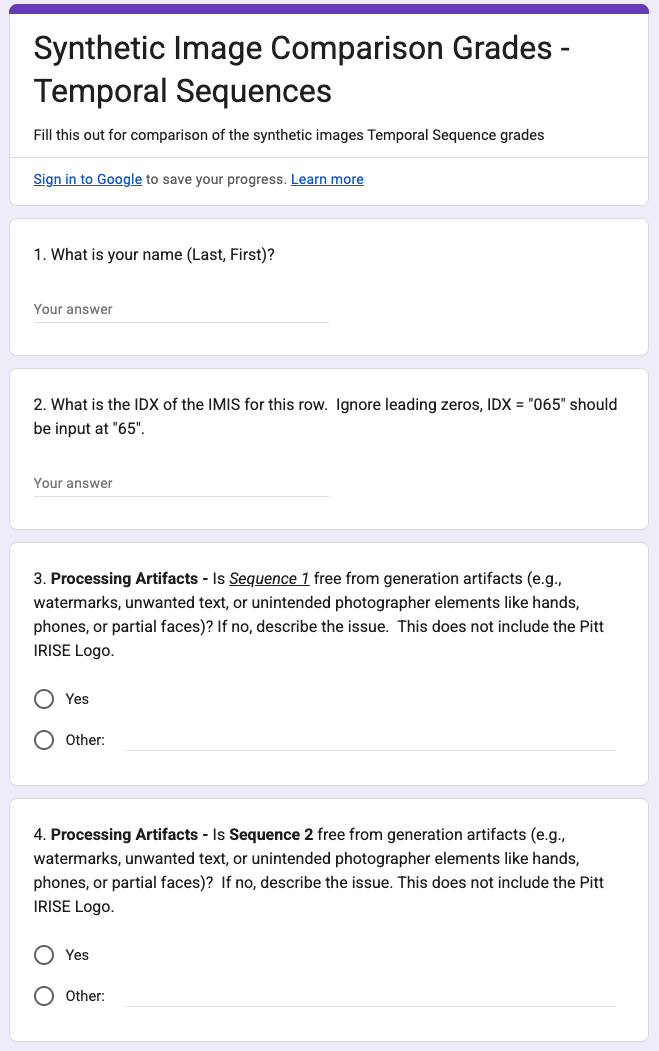}
    \caption{Sample image of the evaluation platform used by reviewers to evaluate the single-pass images and temporal sequences.}
    \label{fig:appendix1}
\end{figure}

Figure~\ref{fig:appendix1} provides a sample screenshot of the Google Form used as the evaluation platform for the expert-based review. Reviewers filled out this form twice for each OSHA narrative reviewed. The form instructs reviewers to simultaneously evaluate both images/sequences created for that OSHA narrative under that methodology. 

\section{Text and Image Example}
This section walks through the text and image generation of one OSHA narrative. This narrative was chosen because it produced both highly rated and poorly rated images during the expert review. 

The fields most relevant to image generation for this narrative are reproduced below:
\begin{center}
\small
\renewcommand{\arraystretch}{1.3}
\begin{tabular}{p{0.22\linewidth} p{0.70\linewidth}}
\toprule
\textbf{Field} & \textbf{Value} \\
\midrule
\texttt{event\_desc}     & Employee Killed When Struck by Motor Vehicle \\
\texttt{event\_type}     & Struck By \\
\texttt{src\_of\_injury} & Windstorm/Lightning, etc.\ [administrative classification; hazard is vehicle intrusion] \\
\texttt{event\_keyword}  & HEAD, HIGH WIND, TRAFFIC ACCIDENT, CONSTRUCTION, COLLISION, STRUCK BY, BRIDGE, MOTOR VEHICLE \\
\texttt{degree\_of\_inj} & Fatal \\
\texttt{abstract\_text}  & \textit{``Employee \#1, a highway bridge construction worker, was walking in a bar ditch that was approximately 20 to 25 feet from the shoulder of the road.  A collision between two motor vehicles resulted in one of the vehicles departing the paved surface and striking and killing Employee \#1 as he was walking by.''} \\
\bottomrule
\end{tabular}
\end{center}

\subsection*{Mode I: Single-Pass Image Generation}

The full record was passed to the text generator $f_\theta$ with the structured prompt $P_{SP}$, which instructs the model to define three narrative layers (infrastructure, activity, hazard) in 1--2~sentences each. The output $R_{SP}$ is reproduced verbatim below:

\medskip
\noindent
\textbf{Infrastructure layer.}
\begin{quote}
\small\itshape
``THE INFRASTRUCTURE: A multi-lane highway intersection, N.\ Loop 289 and
Akron Access Road, features a paved surface bordered by a clear shoulder.
Beyond the shoulder, a distinct bar ditch runs parallel to the roadway,
approximately 20 to 25~feet from the paved edge.''
\end{quote}
 
\noindent
\textbf{Activity layer.}
\begin{quote}
\small\itshape
``THE ACTIVITY: A highway bridge construction worker is performing a usual
task, walking within the bar ditch.  The worker maintains a position
approximately 20 to 25~feet from the paved shoulder of the road.''
\end{quote}
 
\noindent
\textbf{Hazard layer.}
\begin{quote}
\small\itshape
``THE HAZARD: On the adjacent paved surface, two motor vehicles collide, causing one vehicle to veer sharply from the roadway.  This vehicle departs the paved surface, crosses the shoulder, and strikes the worker in the head within the bar ditch.''
\end{quote}
 
\begin{figure}[H]
    \centering
    \includegraphics[width=0.9\textwidth]{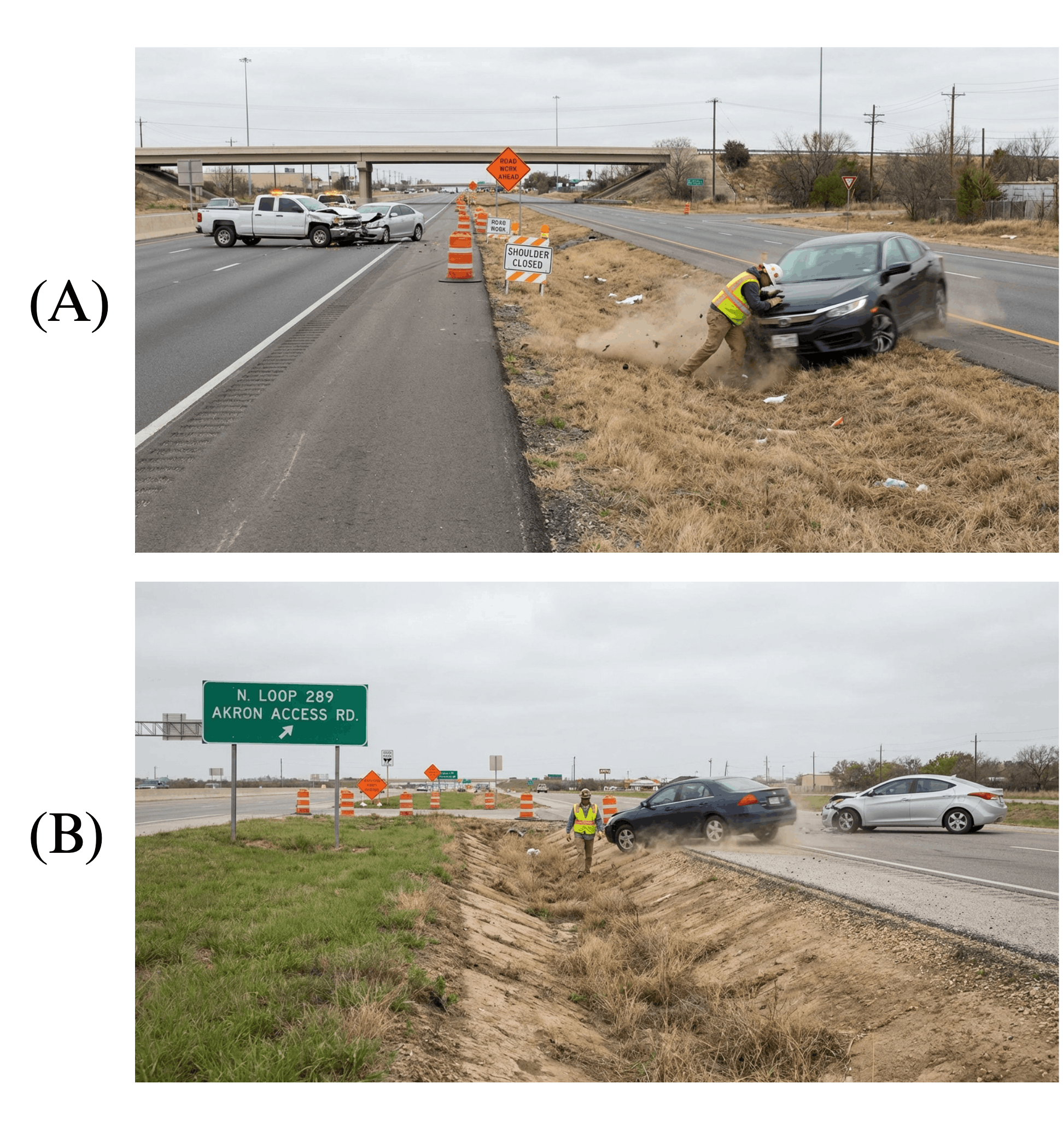}
    \caption{Single-pass outputs of the example OSHA narrative. (A) Iteration 1 and (B) Iteration 2.}
    \label{fig:appendix2}
\end{figure}

The three layers are concatenated to form the complete description $R_{SP}$, which is then embedded in the visual constraints prompt $V_{SP}$. The single-pass image $I_{SP}$ is produced twice using $f_G$ and visualized in Figure~\ref{fig:appendix2}. Because $f_G$ is a stochastic generator, two statistically independent images from the same prompt were obtained. For Iteration 1 (see Figure~\ref{fig:appendix2} (A)), the image received unanimous reviews of "Major issues -- not acceptable". This was primarily because the two car accident is visible on the left side of the image, yet the car hitting the worker is clearly coming from the right side of the image. However, Iteration 2 (see Figure~\ref{fig:appendix2} (B)) clearly depicts the vehicle collision, with one car veering into the median moments away from striking the worker.

\subsection*{Mode II: Temporal Sequences}
 
The temporal pipeline decomposed the OSHA narrative into four sequentially generated text descriptions $\{R_{T1}, R_{T2}, R_{T3}, R_{T4}\}$, each produced by
$f_\theta$ conditioned on the prior output, which then drive a sequential image-to-image conditioning pipeline through $f_G$.  As with the single-pass case, two statistically independent iterations of the full pipeline were executed from identical prompts.
 
\subsubsection*{Stage 1 --- Base Environment ($R_{T1}$)}
 
The first description captures the static highway infrastructure with no workers or equipment present:
 
\begin{quote}
\small\itshape
``The site features a wide, multi-lane paved highway, likely part of Loop~289, designed for significant traffic flow.  A distinct shoulder borders the main roadway, giving way to a broad, unpaved bar ditch that stretches approximately 20 to 25~feet from the road's edge.  Permanent lane markings delineate the travel lanes, and the infrastructure suggests the presence of an existing or planned bridge structure over or under the main thoroughfare.''
\end{quote}
 
\subsubsection*{Stage 2 --- Construction Activity ($R_{T2}$)}
 
Conditioned on $R_{T1}$ and $D$, the activity description $R_{T2} = f_\theta(P_{T2}(D,\, R_{T1}))$ introduces workers and equipment into the established scene:
 
\begin{quote}
\small\itshape
``A large excavator is positioned near the edge of the bar ditch, its bucket resting on a pile of excavated earth, indicating recent or ongoing trenching work for drainage or utility lines.  Further down the ditch, a compact track loader is preparing to grade a section of the unpaved surface, while a dump truck idles on a temporary access path off the shoulder, ready for material transport.  A highway bridge construction worker, wearing a hard hat and a high-visibility vest, is routinely walking along the base of the broad bar ditch, approximately 20~feet from the paved shoulder, inspecting the ground conditions.''
\end{quote}
 
\subsubsection*{Stage 3 --- Safety Warning Annotation ($R_{T3}$)}
 
The warning specification $R_{T3} = f_\theta(P_{T3}(D,\, R_{T2}))$ branches from $R_{T2}$ independently of $R_{T4}$, identifying the at-risk worker and supplying a concise warning phrase:
 
\begin{quote}
\small\itshape
``The highway bridge construction worker, walking along the base of the broad bar ditch approximately 20~feet from the paved shoulder, should be highlighted with a red outline or glow.\\ 
\textsc{safety\_warning}: Watch for vehicles leaving roadway''
\end{quote}
 
\subsubsection*{Stage 4 --- Hazard Event ($R_{T4}$)}
 
The hazard description $R_{T4} = f_\theta(P_{T4}(D,\, R_{T2}))$ depicts the peak worker--hazard interaction:
 
\begin{quote}
\small\itshape
``On the adjacent paved shoulder, a sudden collision occurs between two motor vehicles.  One of the vehicles, now uncontrolled, departs the paved surface and travels into the broad bar ditch.  The motor vehicle then strikes the highway bridge construction worker, who is walking along the ditch base, resulting in a fatal impact to the head.''
\end{quote}
 
\begin{figure}[H]
    \centering
    \includegraphics[width=0.9\textwidth]{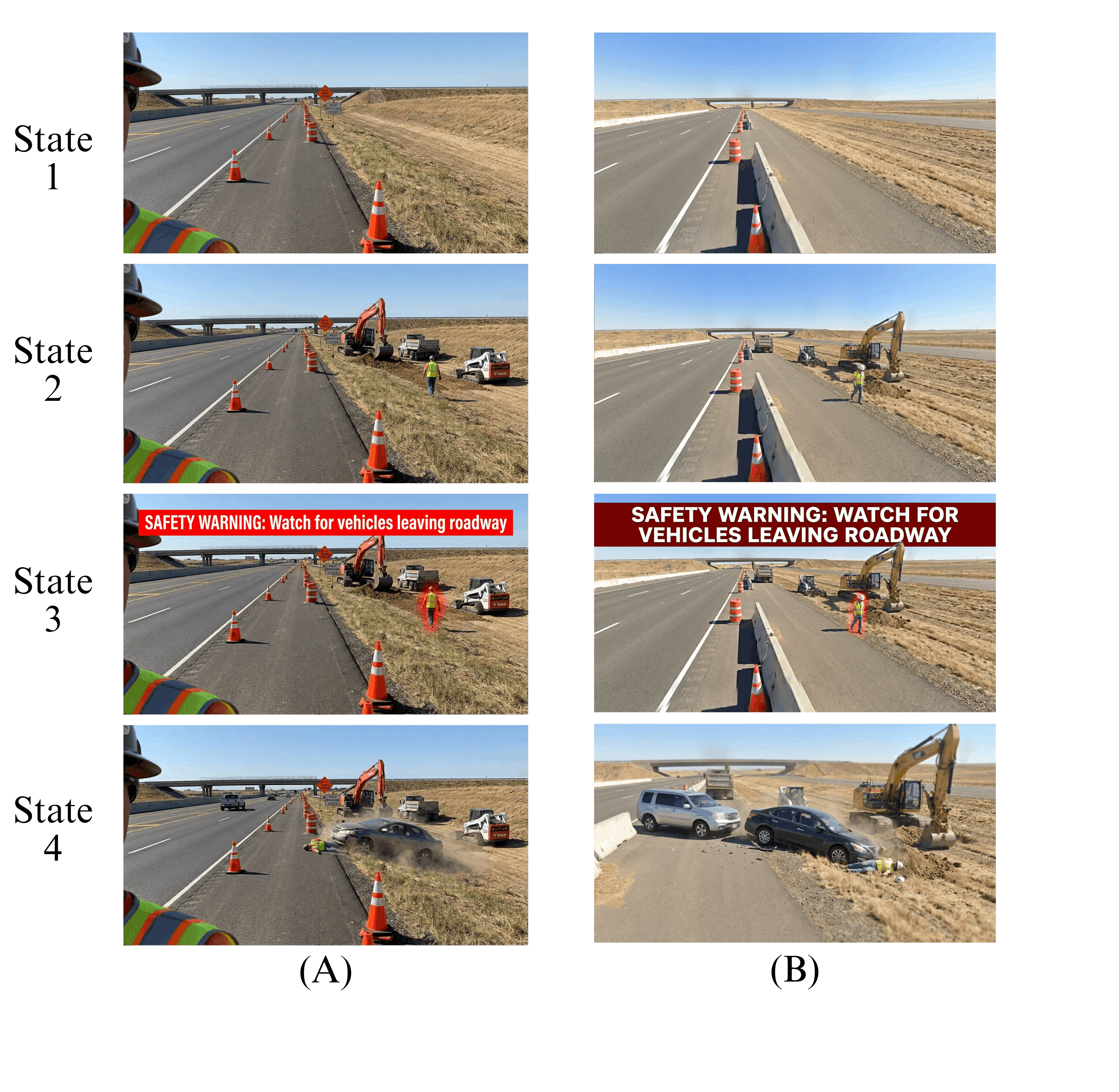}
    \caption{Temporal sequence outputs of the example OSHA narrative.
             (A) Iteration 1 and (B) Iteration 2.  Each row shows the four
             frames: Stages 1, 2, 3, and 4.}
    \label{fig:appendix3}
\end{figure}

\noindent
Each description is paired with its corresponding visual constraints prompt $V_{Ti}$ and passed to $f_G$ to produce images $\{I_{T1}, I_{T2}, I_{T3}, I_{T4}\}$, which are visualized in Figure~\ref{fig:appendix3}.  Sequence~1 (see Figure~\ref{fig:appendix3} (A)) received mixed reviews, primarily because the impacting vehicle approaches from an implausible angle inconsistent with the two-vehicle collision described in $R_{T4}$.  One reviewer also flagged a processing artifact in $I_{T1}$: a partial photographer figure visible on the left side of the frame, which persisted through the remaining stages due to the sequential image-to-image conditioning.  Sequence~2 (see Figure~\ref{fig:appendix3} (B)) received unanimous ``No Issues --- Fully Acceptable'' ratings, with both reviewers awarding 5/5 for fidelity and alignment, citing a clear and accurate depiction of the collision event.
 
\section{Additional Results}

This section reports the full distributional statistics underlying the CLIP retrieval analysis (Table~\ref{tab:appendix-clip-dist}) and the supporting expert-evaluation data, including per-dimension inter-rater agreement, the complete educational utility breakdown, and the stochastic recovery analysis (Table~\ref{tab:appendix-expert}).

\begin{table}[H]
    \centering
    \small
    \caption{Distributional separation, Cohen's $d$, and Manning-Whitney $p$ between matched and mismatched cosine similarities in CLIP space.}
    \label{tab:appendix-clip-dist}
    \begin{tabular}{l cc cc ccc}
        \toprule
        & \multicolumn{2}{c}{\textbf{Mismatched}}
        & \multicolumn{2}{c}{\textbf{Matched}} \\
        \cmidrule(lr){2-3} \cmidrule(lr){4-5}
        \textbf{Group}
        & $n$ & $\mu \pm \sigma$
        & $n$ & $\mu \pm \sigma$
        & \textbf{Sep.}
        & \textbf{Cohen's $d$}
        & \textbf{MW $p$} \\
        \midrule
        Overall
        & 22{,}200 & $0.178 \pm 0.044$
        & 300      & $0.243 \pm 0.040$
        & $1.48\sigma$ & 1.477 & $1.22 \times 10^{-106}$ \\
        Single-pass
        & 11{,}100 & $0.176 \pm 0.043$
        & 150      & $0.249 \pm 0.037$
        & $1.69\sigma$ & 1.692 & $1.81 \times 10^{-65}$ \\
        Temporal
        & 11{,}100 & $0.180 \pm 0.045$
        & 150      & $0.237 \pm 0.043$
        & $1.27\sigma$ & 1.274 & $4.88 \times 10^{-44}$ \\
        \bottomrule
    \end{tabular}
\end{table}

\begin{table}[H]
    \centering
    \small
    \caption{Supporting expert-evaluation data. (A) Inter-rater agreement (Fleiss' $\kappa$) by checklist dimension, with bootstrap 95\% confidence intervals and agreement-strength labels. (B) Full educational utility breakdown, (C) Stochastic recovery from paired comparative evaluations.}
    \label{tab:appendix-expert}

    \textbf{(A) Inter-rater agreement (Fleiss' $\kappa$)}\\[4pt]
    \begin{tabular}{l cc cc}
        \toprule
        & \multicolumn{2}{c}{\textbf{$\kappa$ [95\% CI]}}
        & \multicolumn{2}{c}{\textbf{Agreement}} \\
        \cmidrule(lr){2-3} \cmidrule(lr){4-5}
        \textbf{Dimension}
        & \textbf{Single-Pass} & \textbf{Temporal}
        & \textbf{Single-Pass} & \textbf{Temporal} \\
        \midrule
        Processing Artifacts  & 0.586 [0.219, 0.874] & 0.767 [0.609, 0.887] & Moderate & Substantial \\
        Hazard Accuracy       & 0.343 [0.178, 0.492] & 0.411 [0.233, 0.576] & Fair     & Moderate \\
        Scene Realism         & 0.224 [0.039, 0.394] & 0.195 [0.015, 0.365] & Fair     & Slight \\
        Visual Coherence      & 0.247 [0.075, 0.396] & 0.287 [0.099, 0.459] & Fair     & Fair \\
        Temporal Consistency  & ---                  & 0.494 [0.341, 0.628] & ---      & Moderate \\
        Hazard Alert Accuracy & ---                  & 0.430 [0.228, 0.619] & ---      & Moderate \\
        \bottomrule
    \end{tabular}

    \vspace{10pt}

    \textbf{(B) Educational utility: full distribution and checklist failure rates}\\[4pt]
    \begin{tabular}{l ccc ccc}
        \toprule
        & \multicolumn{3}{c}{\textbf{Single-Pass (\%)}}
        & \multicolumn{3}{c}{\textbf{Temporal (\%)}} \\
        \cmidrule(lr){2-4} \cmidrule(lr){5-7}
        \textbf{Dimension}
        & \textbf{Fully} & \textbf{Minor} & \textbf{Major}
        & \textbf{Fully} & \textbf{Minor} & \textbf{Major} \\
        \midrule
        Processing Artifacts   &  1.9 &  6.1 & 11.1 &  5.4 & 20.2 & 19.7 \\
        Hazard Accuracy        &  0.6 & 57.9 & 93.7 &  4.3 & 26.3 & 75.8 \\
        Scene Realism          &  7.0 & 31.6 & 34.9 & 23.7 & 56.1 & 51.5 \\
        Visual Coherence       &  2.5 & 23.7 & 15.9 &  6.5 & 15.8 & 28.0 \\
        Temporal Consistency   & ---  & ---  & ---  &  8.6 & 40.4 & 61.4 \\
        Hazard Alert Accuracy  & ---  & ---  & ---  &  2.2 &  8.8 & 25.8 \\
        \midrule
        \textit{Rating share}  & 47.0 & 34.1 & 18.9 & 27.4 & 33.6 & 38.9 \\
        \bottomrule
    \end{tabular}

    \vspace{10pt}

    \textbf{(C) Stochastic recovery from paired evaluations}\\[4pt]
    \begin{tabular}{l cc}
        \toprule
        \textbf{Outcome}
        & \textbf{Single-Pass (\%)}
        & \textbf{Temporal (\%)} \\
        \midrule
        Both acceptable       & 70.1 & 44.4 \\
        Mixed                 & 22.2 & 33.1 \\
        Both rejected         &  7.8 & 22.5 \\
        \midrule
        \textbf{Recovery rate} & 74.0 [60.5, 86.7] & 59.6 [48.6, 71.0] \\
        \bottomrule
    \end{tabular}
\end{table}

\section{Image generation cost}
The cost to create this dataset, as of February 2026, was \$0.30 per million tokens up and \$2.50 per million tokens down for Gemini 2.5 flash, while Gemini 3 Pro costs \$2.00 per million tokens up and \$12.00 per million tokens down, in addition to a \$0.134 baseline fee for each high-resolution image generation.

\end{document}